\pdfoutput=1
\documentclass[11pt]{article}
\usepackage[preprint]{acl}
\usepackage{times}
\usepackage{latexsym}
\usepackage[T1]{fontenc}
\usepackage[utf8]{inputenc}
\usepackage{microtype}
\usepackage{booktabs}
\usepackage{multirow}
\usepackage{graphicx}
\usepackage{tabularx}
\usepackage{lscape}
\usepackage{xcolor}
\usepackage{colortbl}
\usepackage{amsmath}
\usepackage{longtable}
\usepackage{tikz}
\usepackage{enumitem}  
\usepackage[normalem]{ulem}
\useunder{\uline}{\ul}{}

\newcommand{\exampleInput}[2]{\small\textbf{#1}: \textit{#2}\\}
\newcommand{\exampleOutput}[2]{\small\textbf{#1}: #2\\}
\newcommand{\addMethod}[1]{~~\textsl{+ #1}}

\definecolor{Gray}{gray}{0.95}

\newcommand{\done}[0]{\textbf{\textcolor{red}{}}}

\newcommand{\ours}[0]{\textsc{TimeBench}}

\title{\textsc{TimeBench}: A Comprehensive Evaluation of Temporal Reasoning Abilities in Large Language Models}

\author{
    Zheng Chu$^{1}$, 
    Jingchang Chen$^{1}$, 
    Qianglong Chen$^{2}$,\\
    \textbf{
    Weijiang Yu$^{3}$, Haotian Wang$^{1}$, Ming Liu$^{1,4}$\thanks{$\,$Corresponding Author.}, Bing Qin$^{1,4}$
    } \\
    $^{1}$Harbin Institute of Technology, Harbin, China \\
    $^{2}$Zhejiang University~~
    $^{3}$Sun Yat-sen University~~
    $^{4}$Peng Cheng Laboratory~~ \\
    \texttt{\{zchu,jcchen,mliu,qinb\}@ir.hit.edu.cn } \\
    \texttt{\{chenqianglong.ai, wanght1998, weijiangyu8\}@gmail.com}
}

\definecolor{bluegray}{HTML}{647D87}

\newcommand{\examplefigure}[2]{
\begin{tikzpicture}[
    every node/.style={outer sep=0},
    window/.style={rectangle, draw=black, rounded corners=2pt, align=left, font=\small, inner xsep=0pt, inner ysep=2mm},
    titlestyle/.style={text=white, font=\bfseries\small},
]
\node[window] (window) at (0,-0.2cm) [minimum width=\linewidth, text width=\linewidth-3mm, anchor=north] {#2};
\draw[draw=black, fill=bluegray] (-\linewidth/2,-0.25cm)
    [sharp corners] -- (\linewidth/2, -0.25cm)
    [rounded corners=2pt] -- (\linewidth/2, 0.25cm)
    [rounded corners=2pt] -- (-\linewidth/2, 0.25cm)
    [sharp corners] -- cycle;
\node[titlestyle] (title) at (0,-0.2mm) [minimum width=\linewidth, text width=\linewidth-4mm, align=left] {\textsc{#1}};
\end{tikzpicture}%
}

\newcommand{\demonstrationfigure}[2]{
\begin{tikzpicture}[
    every node/.style={outer sep=0},
    window/.style={rectangle, draw=black, rounded corners, thick, align=left, font=\small, inner xsep=0pt, inner ysep=3mm},
    titlestyle/.style={text=white, font=\bfseries},
]
\node[window] (window) at (0,-0.2cm) [minimum width=\linewidth, text width=\linewidth-4mm, anchor=north] {#2};
\draw[draw=black, fill=bluegray, thick] (-\linewidth/2,-0.3cm)
    [sharp corners] -- (\linewidth/2, -0.3cm)
    [rounded corners=3pt] -- (\linewidth/2, 0.25cm)
    [rounded corners=3pt] -- (-\linewidth/2, 0.25cm)
    [sharp corners] -- cycle;
\node[titlestyle] (title) at (0,-0.3mm) {#1};
\end{tikzpicture}%
}

\begin{document}
\maketitle

\begin{abstract}
Grasping the concept of time is a fundamental facet of human cognition, indispensable for truly comprehending the intricacies of the world.
Previous studies typically focus on specific aspects of time, lacking a comprehensive temporal reasoning benchmark.
To address this, we propose \ours{}, a comprehensive hierarchical temporal reasoning benchmark that covers a broad spectrum of temporal reasoning phenomena.
\ours{} provides a thorough evaluation for investigating the temporal reasoning capabilities of large language models.
We conduct extensive experiments on GPT-4, LLaMA2, and other popular LLMs under various settings.
Our experimental results indicate a significant performance gap between the state-of-the-art LLMs and humans, highlighting that there is still a considerable distance to cover in temporal reasoning.
Besides, LLMs exhibit capability discrepancies across different reasoning categories.
Furthermore, we thoroughly analyze the impact of multiple aspects on temporal reasoning and emphasize the associated challenges.
We aspire for \ours{} to serve as a comprehensive benchmark, fostering research in temporal reasoning\footnote{$\,$Data is available at: \href{https://github.com/zchuz/TimeBench}{GitHub}}. 

\end{abstract}
\section{Introduction}
\label{sec:introduction}

\textbf{Time flies over us, but leaves its shadow behind.
}
Understanding time is a crucial part of human comprehension of the world.
Envision the blossoming of flowers, and you'll associate it with the arrival of spring.
The ponder within it encompasses the intricate interplay of world knowledge, causality, and event temporal relationships.
Temporal reasoning, in contrast to reasoning of a singular nature, comes with inherent complexity, encompassing implicit arithmetic, logical implications, and world knowledge. It is a form of integrated reasoning built upon foundational reasoning like mathematical and logical reasoning~\citep{gsm8k,lila,reclor}.
Recently, large language models (LLMs) have demonstrated remarkable performance in complex reasoning~\citep{mmlu,bigbench,gpt3,palm,gpt4,llama2}, 
but their performance in temporal reasoning has not yet been extensively explored.

\begin{figure}[t]   
    \centering
    \includegraphics[clip, width=\linewidth]{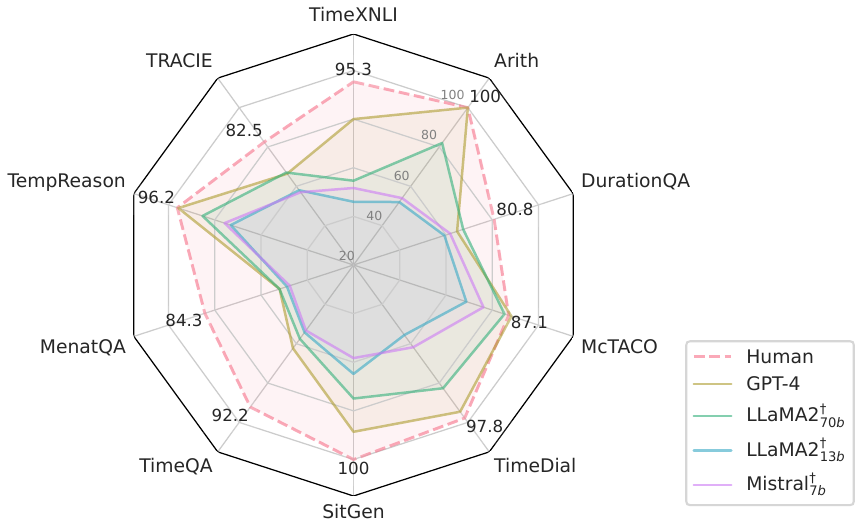}
    \caption{\label{figure:radar}
    A brief overview of human and LLMs' performance on TimeBench. Human scores are annotated.}
\end{figure}

Recent research for temporal reasoning typically focuses only on a few aspects, such as temporal commonsense or temporal question answering~\citep{mctaco,timeqa,templama,tram}. 
Due to the inherent complexity of temporal reasoning, it is challenging to accurately measure models' temporal reasoning capabilities based on limited aspects.

To address this issue, we propose \ours{}, a comprehensive and hierarchical temporal reasoning benchmark.
Specifically, drawing inspiration from the human cognitive process of transitioning from abstraction and concreteness to integration~\citep{psychology}, we categorize temporal reasoning into three levels: symbolic temporal reasoning, commonsense temporal reasoning, and event temporal reasoning. 
These levels respectively represent understanding abstract time expression, grasping concrete world knowledge, and integrating and applying this knowledge in real-world scenarios.
\ours{} comprises 10 tasks with 16 sub-tasks, covering a broad spectrum of temporal reasoning phenomena.
Besides, prior work typically features only a single task form, too simplistic to capture the model's performance.
In contrast, we incorporate four distinct task forms, offering a more realistic simulation of challenges.

To quantify the temporal reasoning capabilities of contemporary LLMs, we extensively assess widely-used LLMs, including proprietary models such as ChatGPT~\citep{gpt3.5} and GPT-4~\citep{gpt4}, as well as open-source like LLaMA2~\citep{llama2}, Vicuna-1.5~\citep{vicuna:chiang}, Mistral~\citep{mistral}, Baichuan2~\citep{baichuan2}, ChatGLM3~\cite{chatglm} and FLAN-T5~\citep{flant5}.
We conduct experiments under zero-shot and few-shot settings, combining commonly used reasoning techniques, chain-of-thought prompting~\citep{zeroshotcot, fewshotcot}.
The experimental results suggest that GPT-4 outperforms other models, showcasing strong temporal reasoning capabilities, as shown in Figure~\ref{figure:radar}. 
Nevertheless, there is still a considerable gap between the strongest models and humans.
On the contrary, open-source models show inferior performance in temporal reasoning, attributed to shortcomings in abstract time understanding, temporal relations modeling, and a lack of temporal commonsense.
In addition, we also observe that chain-of-thought prompting does not yield a consistent improvement in performance.
These findings indicate that there is still significant room for improvement in models' temporal reasoning capabilities.
Moreover, we have conducted a thorough analysis of the deficiencies and obstacles faced by models in temporal reasoning.

We aspire for temporal reasoning to garner increased attention within the research community.
Our contributions can be summarized as follows:
\begin{itemize}[itemsep=2pt,topsep=2pt,parsep=2pt,leftmargin=*]
    \item We introduce \ours{}, a comprehensive and hierarchical benchmark to quantify the temporal reasoning abilities of LLMs.
    \item We conduct extensive experiments with several LLMs, revealing a significant gap between even SOTA LLM and humans, indicating substantial research opportunities in this field.
    \item By conducting a thorough analysis, we reveal the dilemmas that LLMs face in temporal reasoning and identify potential solutions.
\end{itemize}

\section{\textsc{TimeBench} Benchmark}
\label{sec:benchmark}

\subsection{Benchmark Design Principal\done}

\textsc{TimeBench} focuses on a comprehensive evaluation of the temporal reasoning capabilities of large language models in challenging and complex scenarios. 
To achieve this goal, we summarize the difficulties and challenges faced in temporal reasoning, categorize them into three levels, 
and integrate diverse task formats to better align with the intricate nature of temporal reasoning.

Just as the human cognitive process unfolds from foundational cognition and conceptual understanding to practical reasoning, we delineate temporal reasoning into three hierarchical levels.
Specifically, \ours{} categorizes temporal reasoning into symbolic, commonsense and event temporal reasoning, covering 10 datasets with a total of 16 subtasks.
(1) Symbolic Temporal Reasoning focuses on the comprehension of fundamental abstract temporal expressions.
(2) Temporal Commonsense Reasoning emphasizes the mastery of temporal principles, concepts and world knowledge.
(3) Event Temporal Reasoning concentrates on modeling the temporal relationships between events and times within authentic scenarios.

\subsection{Difficulties and Challenges}

We delineate the essential competencies and the challenges that arise from a human cognitive standpoint in the realm of temporal reasoning, and language models confront similar challenges.
We present the dataset statistics, task formats, and the associated challenges in Table~\ref{tab:dataset_challenges}.

\paragraph{Time Expression Understanding}
Time expressions (TimeX) denote words or phrases that convey information about time and represent the simplest and most basic units of expressing time, such as \textit{in April 2000}, \textit{after 2008}.
Grasping time expressions is the most foundational step in understanding temporal elements within the textual modality.

\paragraph{Temporal Commonsense}
assesses the understanding of temporal world knowledge, including \text{event order}, \text{event duration}, \text{typical time}, \text{event frequency} and \text{stationary}, which is crucial for language models to comprehend daily scenarios.

\paragraph{Event-Time Relations} assesses the model's grounding capability to establish temporal relationships between events and their temporal context,
thereby enabling models to grasp the progression and transformations of events as they dynamically evolve through time.

\paragraph{Event-Event Relations} 
not only involve event-time grounding but also introduce multi-hop relative connections between events.
Models with this capability can better handle temporal reasoning in complex scenarios involving multiple events.

\paragraph{Implicit Temporal Reasoning} involves going beyond the surface of texts, engaging in deeper reasoning such as drawing upon temporal commonsense, identifying implicit temporal factors and discerning hidden temporal relationships among events.
Implicit temporal reasoning is pivotal in complex real-world scenarios where events and time are intricately interwoven.

\begin{table}[t]
\centering
\examplefigure{Date Arith}{
    \exampleInput{Q}{What is the time 2 year and 4 month before Mar, 1755}
    \exampleOutput{A}{Nov, 1752}
}
\examplefigure{TimeX NLI}{
    \exampleInput{Premise}{On 28th May 1967, I graduated.}
    \exampleInput{Hypothesis}{Before 23rd October 1920, I graduated.}
    \exampleOutput{A}{Contradiction}
}

\caption{Examples of symbolic temporal reasoning}
\label{tab:symbolic_reasoning_example}
\end{table}

\begin{table}[t]
\centering
\examplefigure{MCTACO}{
    \exampleInput{C}{Ransome looks after her as well as for young Fern Simon , who has declared her love for him.}
    \exampleInput{Q}{How often do Ransome and Fern talk?}
    \exampleOutput{O}{each century, \underline{once a day}, once a century, \underline{every night}}
}
\examplefigure{TimeDial}{
    \exampleInput{Dialog}{... Person1: Do you go to work by train every day \quad Person2: Yes . I commute <MASK> a week by train...}
    \exampleOutput{O}{\underline{five days}, 25 days, a minute, \underline{six days}}
}
\examplefigure{SituatedGen}{
    \exampleInput{Keywords}{axis, one day, one month, Earth, Moon}
    \exampleOutput{A}{\underline{Earth} rotates on its \underline{axis} once in \underline{one day}. It takes \underline{one month} for the \underline{Moon} to rotate on its \underline{axis}.}
}

\caption{Examples of commonsense temporal reasoning.}
\label{tab:commonsense_reasoning_example}
\end{table}

\subsection{Symbolic Temporal Reasoning\done}

To evaluate the language model's comprehension of abstract time expressions, we utilize two symbolic reasoning tasks stripped of semantic content: date arithmetic and time expression inference.
Table~\ref{tab:symbolic_reasoning_example} shows examples of symbolic temporal reasoning.

\paragraph{Date Arithmetic}\citep{tempreason} assesses the model's grasp of abstract date calculation. 
When provided with a date, the model needs to accurately calculate the date a certain amount of time before or after the given date. The smallest unit is one day.

\paragraph{TimeX NLI}\citep{timexnli} 
focuses on the logical entailment relationships among abstract TimeX, including three aspects: order (s1), duration (s2), and duration with unit conversion (s3).

\begin{table}[t]
\centering
\examplefigure{TimeQA}{
    \exampleInput{C}{... He worked in Utrecht for the firm of P Smits \& de Wolf from 1864 to 1867 and then returned to ...}
    \exampleInput{Q}{Where did Ludwig Mond work between Mar 1866 and Sep 1866?}
    \exampleOutput{A}{Utrecht}
}
\examplefigure{MenatQA}{
    \exampleInput{C}{... After the French evacuated Egypt in 1801, Hurshid Pasha was named governor of Egypt in 1804.  Muhammad Ali had himself named governor of Egypt in May 1805 ...}
    \exampleInput{Q}{Which position did Hurshid Pasha hold from 1804 to 1806, if Hurshid Pasha tepped down as the governor of Egypt in 1808?}
    \exampleOutput{A}{ governor of Egypt}
}
\examplefigure{TempReason}{
    \exampleInput{C}{ ... Peter Corke works for Queensland University of Technology from Jan, 2010 to Dec, 2022. Peter Corke works for Commonwealth Scientific from Jan, 1984 to Jan, 2009. ...}
    \exampleInput{Q}{Which employer did Peter Corke work for before Queensland University of Technology?}
    \exampleOutput{A}{Commonwealth Scientific}
}

\caption{Examples of event temporal reasoning.}
\label{tab:event_temporal_reasoning_example}
\end{table}

\subsection{Commonsense Temporal Reasoning\done}
We measure the model's mastery of temporal commonsense and world knowledge, along with its capacity for reasoning based on these insights.
Table~\ref{tab:commonsense_reasoning_example} presents examples of temporal commonsense reasoning in QA and generation forms.

\paragraph{MCTACO}\citep{mctaco} evaluates diverse commonsense knowledge from different aspects of events, including duration, frequency, order, stationary and typical event time.

\paragraph{DurationQA}\citep{durationqa} focuses specifically on temporal commonsense reasoning in the spectrum of event duration.

\paragraph{TimeDial}\citep{timedial} considers temporal commonsense reasoning in dialogue scenarios and involves various aspects of commonsense associated with duration, order, and world knowledge.

\paragraph{SituatedGen}\citep{situatedgen} considers generative commonsense reasoning in a constrained text generation scenario.
Given a set of contrasting keywords, the model needs to choose appropriate keywords for each sentence and generate a pair of contrasting sentences that satisfy temporal commonsense.

\subsection{Event Temporal Reasoning\done}
Event temporal reasoning assesses the model's understanding of relationships between events and time in real-world scenarios, as well as its ability to reasoning under certain temporal or event constraints.
Examples are shown in Table~\ref{tab:event_temporal_reasoning_example}.

\paragraph{TimeQA}\citep{timeqa}
requires the model to answer time-sensitive questions based on context containing numerous time-involved facts.
It is categorized into explicit reasoning and implicit reasoning based on time indicators (before, in, etc.).

\paragraph{MenatQA}\citep{menatqa}
introduces time-sensitive factors to elicit implicit temporal reasoning, including time scope change, disruption of facts, and counterfactual questions, which provides a more in-depth assessment of implicit reasoning ability on event-time relations.

\paragraph{TempReason}\citep{tempreason}
removes irrelevant context and focuses on implicit temporal reasoning within structured facts, investigating the model's capability boundaries. 
It involves event-time reasoning and event-event reasoning.

\paragraph{TRACIE}\citep{tracie} evaluates the model's comprehension of temporal order between implicit events. The model needs to identify events implied in the context and then determine their chronological order.

\subsection{Task Formats and Evaluation Metrics}
\ours{} is a multispectral benchmark encompassing four task types: free-form reading comprehension, natural language inference, constrained text generation, and multi-select questions.
For detailed task types and their corresponding evaluation metrics, please refer to Appendix~\ref{appendix:a:task_formats} and~\ref{appendix:a:evaluation_metrics}.

\section{Methodology\done}
\label{sec:methods}

We perform evaluations using the prompt-based approach, including standard prompting and chain-of-thought prompting. 
Experiments are conducted under both zero-shot and few-shot settings.

\paragraph{Standard Prompting}
We formulate specific instructions for each task.
In the zero-shot setting, models follow the instructions to answer questions. 
In the few-shot setting, models are provided with several question-answer pairs as demonstrations and emulate those instances to answer questions.
\begin{align}
    &\mathrm{prompt_{zs}^{sp}} = \{\mathrm{INST}\}\{\mathrm{Q}\} \\
    &\mathrm{prompt_{fs}^{sp}} = \{\mathrm{INST}\}\{\mathrm{Q_{1}}\}\{\mathrm{A_{1}}\}..\{\mathrm{Q}\}
\end{align}

\paragraph{Chain-of-Thought Prompting}
The instructions of CoT are the same as standard prompting.
In the zero-shot setting, following Zeroshot CoT~\citep{zeroshotcot}, we add a reasoning trigger \textit{Let's think step by step} after questions to perform chain-of-thought reasoning.
In the few-shot setting, we manually annotate CoT demonstrations for each task to guide the step-by-step reasoning.
Prompts can be found in Appendix~\ref{appendix:prompts}.
\begin{align}
    &\mathrm{prompt_{zs}^{cot}} = \{\mathrm{INST}\}\{{\mathrm{Q}}\}\{{\mathrm{TRIG}}\} \\
    &\mathrm{prompt_{fs}^{cot}} = \{\mathrm{INST}\}\{{\mathrm{Q_1}}\}\{{\mathrm{R_1}}\}\{{\mathrm{A_1}}\}..\{{\mathrm{Q}}\}
\end{align}

\section{Experimental Setup\done}
\label{sec:experiments}

\begin{table*}[ht]
\centering
\small
\setlength\tabcolsep{2pt}
\resizebox{\textwidth}{!}{%

\begin{tabular}{lcccccccccccccccccccc}
\toprule

\multicolumn{1}{c}{\multirow{3.5}{*}{\textbf{Method}}} & \multicolumn{4}{c}{\textbf{Symbolic}} & \multicolumn{4}{c}{\textbf{Commonsense}} & \multicolumn{8}{c}{\textbf{Event Temporal}} & \multicolumn{4}{c}{\textbf{Overall}} \\
\cmidrule(lr){2-5}\cmidrule(lr){6-9}\cmidrule(lr){10-17}\cmidrule(lr){18-21}
 & \multicolumn{3}{c}{TimeXNLI} & \multirow{2}{*}{Arith} & \multirow{2}{*}{DQA} &  \multirow{2}{*}{McT.} &  \multirow{2}{*}{TiD.} &  \multirow{2}{*}{SitGen} & \multicolumn{2}{c}{TimeQA} & \multicolumn{3}{c}{MenatQA} & \multicolumn{2}{c}{TempR} &  \multirow{2}{*}{TRACIE} & \multirow{2}{*}{Sym.} & \multirow{2}{*}{Comm.} & \multirow{2}{*}{Event} & \multirow{2}{*}{Avg.} \\
 
 & \textit{s1} & \textit{s2} & \textit{s3} &   &   & &  &   & \textit{Exp.} & \textit{Imp.} & \textit{Sco.} & \textit{Ord.} & \textit{Ctf.} & \textit{L2} & \textit{L3} &  & & & & \\
\midrule

Human & 98.0 & 96.0 & 92.0 & 100.0 & 80.8 & 87.1 & 97.8 & 100.0 & 93.3 & 91.1 & 85.6 & 87.3 & 79.9 & 97.1 & 95.3 & 82.5 & 96.5 & 91.4 & 89.0 & 91.5 \\
\midrule

GPT-4 & 85.3 & 73.3 & 53.3 & 100.0 & \textbf{64.8} & \textbf{88.3} & \textbf{94.6} & \textbf{88.6} & \textbf{73.7} & 51.0 & \textbf{72.4} & \textbf{54.8} & 28.7 & 92.4 & \textbf{95.9} & 62.8 & 78.0 & \textbf{84.1} & \textbf{66.5} & \textbf{73.7} \\
\addMethod{FS CoT} & \textbf{92.0} & \textbf{84.0} & \textbf{64.0} & \textbf{100.0} & 55.1 & 72.3 & 93.4 & - & 66.9 & \textbf{52.8} & 65.3 & 52.6 & 25.9 & \textbf{96.9} & 94.6 & \textbf{66.4} & \textbf{85.0} & 73.6 & 65.2 & 72.1 \\
\midrule

GPT-3.5 & 52.0 & 68.4 & 31.6 & 63.6 & \textbf{67.7} & 71.2 & \textbf{76.4} & \textbf{79.1} & 66.1 & 48.4 & 43.2 & \textbf{51.6} & 17.9 & 84.7 & \textbf{78.0} & 55.0 & 53.9 & \textbf{73.6} & 55.6 & \textbf{59.7} \\
\addMethod{FS CoT} & 51.6 & \textbf{71.8} & 36.6 & \textbf{84.4} & 41.2 & 38.1 & 71.1 & - & \textbf{68.0} & 47.0 & 42.5 & 41.7 & 37.8 & \textbf{89.9} & 76.6 & 50.2 & \textbf{61.1} & 50.1 & \textbf{56.7} & 56.6 \\
\midrule

LLaMA2$_{\text{70b}}^{\dag}$ & \textbf{55.0} & 61.0 & 37.0 & \textbf{82.0} & \textbf{67.4} & \textbf{85.3} & \textbf{82.7} & \textbf{74.9} & \textbf{66.7} & 48.3 & \textbf{61.4} & 42.5 & 33.8 & 85.2 & \textbf{85.4} & 61.0 & 58.8 & \textbf{77.6} & 60.5 & \textbf{64.4} \\
\addMethod{FS CoT} & 52.0 & \textbf{73.0} & \textbf{39.0} & 79.5 & 62.3 & 79.1 & 61.1 & - & 64.3 & 43.0 & 57.7 & 45.2 & \textbf{53.1} & \textbf{87.5} & 81.6 & \textbf{67.0} & \textbf{60.9} & 67.5 & \textbf{62.4} & 63.0 \\
\midrule

LLaMA2$_{\text{13b}}^{\dag}$ & 50.0 & 54.0 & 30.0 & 29.5 & 53.3 & 66.0 & 55.6 & 64.8 & 59.3 & 48.6 & 49.6 & 43.4 & 37.5 & 78.7 & 62.7 & 58.0 & 40.9 & 59.9 & 54.7 & 52.6 \\
\addMethod{FS CoT} & 40.0 & 61.0 & 37.0 & 52.0 & 59.3 & 68.8 & 40.8 & - & 59.4 & 49.1 & \textbf{58.4} & 43.8 & 44.1 & 78.0 & 68.2 & 58.0 & 47.5 & 56.3 & 57.4 & 54.5 \\
\midrule

LLaMA2$_{\text{7b}}^{\dag}$ & 26.0 & 50.0 & 30.0 & 20.0 & 54.5 & 59.6 & 45.2 & 62.4 & 54.4 & 45.3 & 49.8 & 41.9 & 35.8 & 64.0 & 53.3 & 49.0 & 31.5 & 55.4 & 49.2 & 46.3 \\
\addMethod{FS CoT} & 37.0 & 52.0 & 36.0 & 25.5 & 56.9 & 67.0 & 41.9 & - & 45.6 & 36.1 & 50.9 & 38.0 & \textbf{57.3} & 59.7 & 57.7 & 50.0 & 37.6 & 55.3 & 49.4 & 47.4 \\
\midrule

Baichuan2$_{\text{13b}}^{\dag}$ & 38.0 & 48.0 & 33.0 & 42.5 & 54.8 & 73.0 & 45.7 & 64.9 & 59.4 & \textbf{54.2} & 52.7 & 38.0 & 21.4 & 77.3 & 63.5 & 54.0 & 40.4 & 59.6 & 52.6 & 51.3 \\
\addMethod{FS CoT} & 50.0 & 56.0 & 34.0 & 47.0 & 62.0 & 69.3 & 43.8 & - & 58.2 & 49.6 & 49.8 & 40.1 & 45.6 & 81.3 & 65.6 & \textbf{60.0} & 46.8 & 58.4 & 56.3 & 54.2 \\
\midrule

Baichuan2$_{\text{7b}}^{\dag}$ & 27.0 & 66.0 & \textbf{41.0} & 32.5 & 59.8 & 69.4 & 34.3 & 59.8 & 53.8 & \textbf{50.2} & 49.6 & 38.5 & 22.9 & 65.9 & 51.0 & 55.0 & 41.6 & 55.8 & 48.4 & 48.5 \\
\addMethod{FS CoT} & 30.0 & 56.0 & 34.0 & 34.0 & 57.0 & 69.5 & 44.5 & - & 51.2 & 40.7 & 46.4 & 32.6 & \textbf{46.3} & 61.5 & 64.1 & 53.0 & 38.5 & 57.0 & 49.5 & 48.1 \\
\midrule

Mistral$_{\text{7b}}^{\dag}$ & 48.0 & 53.0 & 38.0 & 41.0 & 61.8 & \textbf{76.2} & 61.8 & 58.3 & 55.9 & 45.3 & 49.4 & 47.8 & 45.5 & 76.7 & 74.8 & 53.0 & 45.0 & 64.5 & 56.1 & 55.4 \\
\addMethod{FS CoT} & \textbf{57.0} & 63.0 & 35.0 & 54.0 & 61.8 & 45.7 & 57.3 & - & 60.4 & 46.2 & 57.2 & \textbf{47.9} & 33.2 & 65.9 & 67.9 & 57.0 & 52.3 & 54.9 & 54.5 & 54.0 \\
\midrule

ChatGLM3$_{\text{6b}}^{\dag}$ & 48.0 & 70.0 & 32.0 & 35.0 & 51.8 & 62.6 & 55.0 & 61.6 & 57.2 & 26.3 & 35.4 & 41.5 & 22.5 & 76.4 & 55.9 & 58.0 & 46.3 & 57.8 & 46.7 & 49.3 \\
\addMethod{FS CoT} & 47.0 & 68.0 & 32.0 & 46.0 & 53.9 & 64.3 & 56.5 & - & 52.5 & 24.5 & 35.0 & 40.2 & 22.5 & 79.4 & 60.3 & 54.0 & 48.3 & 58.2 & 46.1 & 49.1\\
\bottomrule
\end{tabular}
}
\caption{
Experimental results under \textbf{few-shot} settings (standard prompting by default). 
$^{\dag}$ denotes the base model without alignment.
Global top-3 results are \textbf{bold}.
Figure~\ref{fig:overall} provides a horizontal comparison of the performance of all models.
Full results in Appendix~\ref{appendix:full_results}.}
\label{tab:fewshot-results}
\end{table*}

\begin{table*}[ht]
\centering
\small
\setlength\tabcolsep{2pt}
\resizebox{\textwidth}{!}{%

\begin{tabular}{lcccccccccccccccccccc}
\toprule

\multicolumn{1}{c}{\multirow{3.5}{*}{\textbf{Method}}} & \multicolumn{4}{c}{\textbf{Symbolic}} & \multicolumn{4}{c}{\textbf{Commonsense}} & \multicolumn{8}{c}{\textbf{Event Temporal}} & \multicolumn{4}{c}{\textbf{Overall}} \\
\cmidrule(lr){2-5}\cmidrule(lr){6-9}\cmidrule(lr){10-17}\cmidrule(lr){18-21}
 & \multicolumn{3}{c}{TimeXNLI} & \multirow{2}{*}{Arith} & \multirow{2}{*}{DQA} &  \multirow{2}{*}{McT.} &  \multirow{2}{*}{TiD.} &  \multirow{2}{*}{SitGen} & \multicolumn{2}{c}{TimeQA} & \multicolumn{3}{c}{MenatQA} & \multicolumn{2}{c}{TempR} &  \multirow{2}{*}{TRACIE} & \multirow{2}{*}{Sym.} & \multirow{2}{*}{Comm.} & \multirow{2}{*}{Event} & \multirow{2}{*}{Avg.} \\
 
 & \textit{s1} & \textit{s2} & \textit{s3} &   &   & &  &   & \textit{Exp.} & \textit{Imp.} & \textit{Sco.} & \textit{Ord.} & \textit{Ctf.} & \textit{L2} & \textit{L3} &  & & & & \\
\midrule

Human & 98.0 & 96.0 & 92.0 & 100.0 & 80.8 & 87.1 & 97.8 & 100.0 & 93.3 & 91.1 & 85.6 & 87.3 & 79.9 & 97.1 & 95.3 & 82.5 & 96.5 & 91.4 & 89.0 & 91.5 \\
\midrule

GPT-4 & 78.6 & 76.0 & 50.7 & \textbf{98.0} & \textbf{59.2} & 80.0 & \textbf{91.1} & \textbf{59.3} & 60.6 & \textbf{46.5} & \textbf{57.0} & 57.0 & 23.1 & 95.3 & \textbf{95.0} & \textbf{64.8} & 75.8 & 72.4 & \textbf{62.4} & 68.3 \\
\addMethod{CoT} & \textbf{80.0} & \textbf{76.0} & \textbf{60.0} & 92.0 & 58.1 & \textbf{82.6} & 89.3 & - & 61.3 & 41.2 & 54.6 & \textbf{59.6} & 22.6 & \textbf{97.0} & 94.5 & 58.0 & \textbf{77.0} & \textbf{76.7} & 61.1 & \textbf{68.5} \\
\midrule

GPT-3.5 & 45.4 & \textbf{67.6} & 31.2 & \textbf{97.0} & 50.5 & \textbf{68.6} & \textbf{69.1} & \textbf{62.3} & \textbf{70.8} & 35.4 & 40.9 & 43.9 & 22.9 & \textbf{81.2} & \textbf{73.8} & 57.4 & \textbf{60.3} & \textbf{62.6} & \textbf{53.3} & \textbf{57.4} \\
\addMethod{CoT} & 33.6 & 64.8 & 33.6 & 71.0 & 23.2 & 45.1 & 67.0 & - & 64.4 & 35.1 & 39.7 & 42.9 & 26.3 & 57.6 & 68.1 & 52.0 & 50.8 & 45.1 & 48.3 & 48.3 \\
\midrule

LLaMA2$_{70b}$ & \textbf{44.0} & 47.0 & 32.0 & \textbf{78.5} & \textbf{59.2} & \textbf{68.9} & 57.0 & 25.0 & 40.8 & \textbf{40.6} & 18.9 & 16.6 & 12.0 & 63.5 & 54.5 & 48.0 & \textbf{50.4} & 52.5 & 36.8 & 44.1 \\
\addMethod{CoT} & 30.0 & 66.0 & 28.0 & 53.5 & 57.3 & 67.1 & \textbf{58.6} &  & 31.4 & 19.5 & 12.2 & 12.7 & 20.8 & 37.5 & 40.5 & 51.0 & 44.4 & \textbf{61.0} & 28.2 & 39.1 \\
\midrule

LLaMA2$_{13b}$ & 30.0 & 49.0 & 34.0 & 22.5 & 38.5 & 40.6 & 35.4 & 57.9 & \textbf{61.9} & 30.5 & 46.1 & 36.1 & 26.9 & 53.1 & 69.4 & 49.0 & 33.9 & 43.1 & 46.6 & 42.6 \\
\addMethod{CoT} & 36.0 & 50.0 & 38.0 & 6.0 & 39.2 & 51.7 & 36.9 & - & 58.7 & 38.9 & 40.9 & 32.5 & \textbf{33.6} & 58.0 & 68.4 & 47.0 & 32.5 & 42.6 & 47.3 & 42.4 \\
\midrule

LLaMA2$_{7b}$ & 39.0 & 53.0 & 30.0 & 13.0 & 39.3 & 41.0 & 6.3 & 24.5 & 49.0 & 29.0 & 26.8 & 21.1 & 16.0 & 63.9 & 47.9 & 49.0 & 33.8 & 27.8 & 37.8 & 34.3 \\
\addMethod{CoT} & 44.0 & 50.0 & 33.0 & 5.0 & 35.0 & 40.0 & 1.7 & - & 49.9 & 31.6 & 31.4 & 24.5 & 17.8 & 56.9 & 48.1 & 46.0 & 33.0 & 25.6 & 38.3 & 34.3 \\
\midrule

Baichuan2$_{13b}$ & 41.0 & \textbf{61.0} & 37.0 & 12.5 & 52.0 & 63.4 & 57.7 & 52.2 & 55.4 & 34.6 & 48.8 & \textbf{44.3} & 39.5 & 57.4 & 61.4 & 49.0 & 37.9 & 56.3 & 48.8 & 48.0 \\
\addMethod{CoT} & 40.0 & 57.0 & 31.0 & 10.0 & 44.6 & 61.9 & 58.1 & - & 41.5 & \textbf{40.9} & \textbf{52.0} & 38.5 & \textbf{43.2} & 62.8 & 64.3 & 55.0 & 34.5 & 54.9 & 49.8 & 46.7 \\
\midrule

Baichuan2$_{7b}$ & 35.0 & 50.0 & 37.0 & 4.5 & 47.9 & 55.3 & 54.3 & 42.0 & 41.5 & 34.7 & 35.2 & 31.2 & 20.4 & 43.4 & 47.7 & 55.0 & 31.6 & 49.9 & 38.6 & 39.7 \\
\addMethod{CoT} & 38.0 & 43.0 & 32.0 & 1.0 & 37.9 & 58.0 & 44.2 & - & 53.5 & 38.8 & 39.9 & 33.2 & 29.3 & 41.2 & 47.2 & 54.0 & 28.5 & 46.7 & 42.1 & 39.4 \\
\midrule

Vicuna1.5$_{13b}$ & 35.0 & 50.0 & 36.0 & 15.0 & 39.2 & 59.1 & 34.2 & 51.8 & 60.4 & 37.0 & \textbf{46.8} & 37.4 & 23.2 & 42.1 & 43.6 & 46.0 & 34.0 & 46.1 & 42.1 & 41.1 \\
\addMethod{CoT} & 42.0 & 51.0 & 37.0 & 3.0 & 29.8 & 50.0 & 33.7 & - & 56.9 & 36.4 & 38.2 & 37.7 & 20.4 & 49.0 & 49.1 & 51.0 & 33.3 & 37.8 & 42.3 & 39.0 \\
\midrule

Vicuna1.5$_{7b}$ & 37.0 & 58.0 & 43.0 & 5.0 & 40.4 & 52.5 & 32.0 & 47.8 & 47.1 & 18.5 & 35.7 & 25.7 & 17.3 & 33.0 & 46.8 & 54.0 & 35.8 & 43.2 & 34.8 & 37.1 \\
\addMethod{CoT} & 36.0 & 50.0 & 36.0 & 1.5 & 39.4 & 49.2 & 36.2 & - & 40.9 & 24.6 & 26.2 & 28.5 & 25.0 & 27.7 & 40.3 & 54.0 & 30.9 & 41.6 & 33.4 & 34.4 \\
\midrule

FLANT5$_{11b}$ & 53.0 & 63.0 & 43.0 & 0.0 & \textbf{52.0} & 65.0 & 47.7 & 49.5 & 61.7 & 26.8 & 33.6 & \textbf{52.2} & 21.8 & \textbf{87.9} & \textbf{83.9} & \textbf{64.0} & 39.8 & 53.6 & \textbf{54.0} & \textbf{50.3} \\
\addMethod{CoT} & \textbf{56.0} & 66.0 & \textbf{45.0} & 0.0 & 49.7 & 63.4 & 42.7 & - & \textbf{64.4} & 28.2 & 41.6 & 50.2 & \textbf{30.6} & 79.5 & 68.9 & 55.0 & 41.8 & 51.9 & 52.3 & 49.4 \\
\midrule

Mistral$_{7b}$ & 47.0 & 50.0 & \textbf{43.0} & 26.5 & 49.8 & 58.8 & 23.2 & \textbf{58.3} & 28.2 & 21.4 & 24.3 & 22.3 & 21.7 & 39.6 & 31.6 & 51.0 & 41.6 & 47.5 & 30.0 & 37.3 \\
\addMethod{CoT} & 38.0 & 56.0 & 35.0 & 16.5 & 36.6 & 49.3 & 19.3 & - & 31.3 & 22.4 & 21.1 & 24.9 & 25.6 & 34.0 & 31.2 & \textbf{61.0} & 36.4 & 35.1 & 31.4 & 33.5 \\
\midrule

ChatGLM3$_{6b}$ & 38.0 & 50.0 & 34.0 & 2.0 & 34.1 & 43.6 & 56.7 & 38.9 & 41.2 & 31.7 & 33.8 & 26.0 & 32.2 & 57.0 & 54.0 & 50.0 & 31.0 & 43.3 & 40.7 & 39.0 \\
\addMethod{CoT} & 27.0 & 49.0 & 37.0 & 0.0 & 24.8 & 37.1 & 44.8 & - & 41.7 & 25.4 & 34.6 & 28.1 & 41.2 & 44.5 & 52.0 & 48.0 & 28.3 & 35.6 & 39.4 & 35.7\\

\bottomrule
\end{tabular}
}
\caption{
Experimental results under \textbf{zero-shot} settings (standart prompting by default). 
All models are alignment models (-chat or -instruct).
Global top-3 results are \textbf{bold}.
}
\label{tab:zeroshot_results}
\end{table*}

\subsection{Models\done}
We evaluate several popular LLMs, including both open-source and proprietary models, with parameter sizes ranging from 6B to 70B.\footnote{~Since OpenAI has never disclosed the scale of ChatGPT series, 6B to 70B here refers to ChatGLM3$_{\text{6B}}$ to LLaMA2$_{\text{70B}}$.}
The complete list of models can be found in Appendix~\ref{appendix:models}.

\subsection{Implementation Details\done}
\label{sec:implementation}
We access proprietary models through Azure API 0613 version.
For open-source models, we deploy them locally through FastAPI.
We set the temperature to 0.0 for greedy decoding in all experiments.
To improve answer extraction accuracy, we prompt models with trigger \textit{Therefore, the answer is} before model outputs to deduce final answers.

\section{Experimental Results}
\label{sec:results}

\subsection{Few-shot Results}
\label{subsec:fewshot}
Table~\ref{tab:fewshot-results} presents the experimental results under few-shot settings.
GPT-4 achieves the best performance across three categories, while LLaMA2$_{\text{70b}}$ and GPT-3.5 rank in the second tier.
However, there remains a substantial gap of 19.4\% between the most powerful LLM and humans.

In symbolic temporal reasoning tasks, GPT-4 demonstrates exceptional performance. 
However, other models exhibit a significant decline in comparison to GPT-4.
In commonsense temporal reasoning tasks, GPT4 lags behind humans by only 8.0\%,
indicating its powerful internal knowledge reservoir. 
With the model scale shrinking, its knowledge reservoir also decreases gradually, leading to a decline in performance.
Notably, there is a significant gap of 25.2\% between LLMs and humans in event temporal reasoning,
which suggests that LLMs encounter major challenges in modeling intricate event-time relationships.

\subsection{Zero-shot Results}
\label{subsec:zeroshot}
Experimental results of alignment models under zero-shot settings are shown in Table~\ref{tab:zeroshot_results}.
In zero-shot settings, GPT-4 and GPT-3.5 rank first and second, respectively, 
and they significantly outperform all open-source models by a large margin.
It is noteworthy that open-source models exhibit a larger performance decline compared to proprietary models when transitioning from few-shot to zero-shot scenarios.
GPT, Baichuan2 and LLaMA2 suffer drops of 5.6\%, 14.6\% and 27.2\%, respectively.
We attribute this performance decline to the quality of alignment.
Restricted by their limited instruction-following capability,
open-source models struggle to fully unleash their performance solely through instructions.
Therefore, few-shot prompting is a better approach for stimulating their temporal reasoning abilities.

\begin{figure}[t]   
    \centering
    \includegraphics[clip, width=\linewidth]{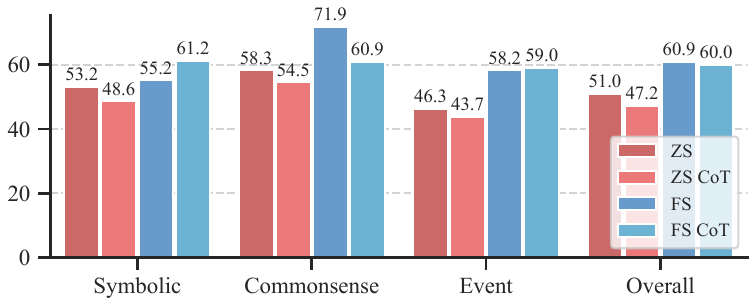}
    \caption{\label{figure:impact_fs_cot}
    Performance gap with and without CoT prompting. 
    The results are averaged from GPT-4, GPT-3.5, Baichuan2$_{13\text{b}}$, LLaMA2$_{70\text{b}}$ and Mistral$_{7\text{b}}$.}
\end{figure}

\subsection{Chain-of-Thought in Temporal Reasoning}
Previous research has found that chain-of-thought prompting can enhance the model's reasoning ability~\citep{fewshotcot, zeroshotcot}.
We aim to explore the following questions: 
\textit{Does CoT prompting bring consistent improvement in temporal reasoning?}
Due to the diversity of temporal reasoning, the above question has not yet been definitively answered.
To investigate this, we select several popular LLMs and analyze their performance affected by chain-of-thought prompting.

\paragraph{Chain-of-thought reasoning is not consistently effective.}
As illustrated in Figure~\ref{figure:impact_fs_cot}, introducing zero-shot CoT prompting results in consistent declines, with an overall decrease of 7.4\%.
In the few-shot scenario, CoT prompting also fails to yield consistent improvements, 
varying depending on the task.
There is a 10.8\% improvement in symbolic reasoning, while a significant decline of 15.2\% in commonsense reasoning.
In event temporal reasoning, there is a slight improvement of 1.3\%.
Next, we will conduct a more detailed analysis of the impact of CoT on specific tasks.

\begin{figure}[t]   
    \centering
    \includegraphics[clip, width=\linewidth]{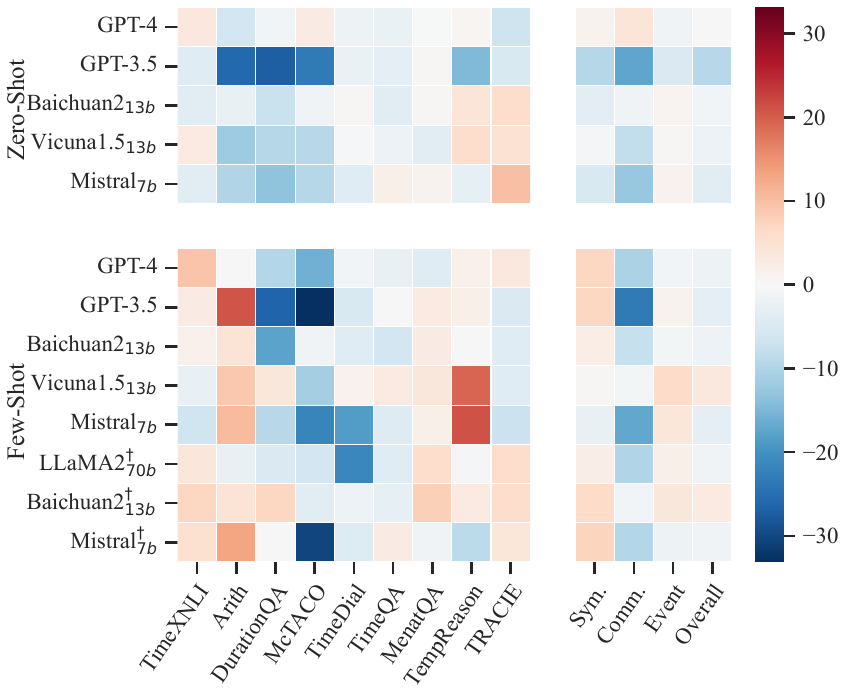}
    \caption{\label{figure:heatmap}
    $\Delta\text{Score}$ between the chain-of-thought prompting and direct I-O prompting. 
    \textbf{Top:} zero-shot setting, \textbf{Bottom:} few-shot setting, \textbf{Left:} variation in each task, \textbf{Right:} averaged variation in the symbolic, commonsense, event, and overall tasks.}
\end{figure}

\paragraph{Impact of CoT prompting across tasks.}

In order to explore the impact of CoT on various tasks thoroughly, we delve into the performance changes of each model across specific tasks within each category, as illustrated in Figure~\ref{figure:heatmap}.
In the zero-shot setting, open-source models achieve a slight improvement in event temporal reasoning with chain-of-thought prompting, while in other cases, they face performance degradation.
While in the few-shot setting, almost all models exhibit significant improvement in symbolic temporal reasoning, with a concurrent prevalent decline in commonsense temporal reasoning.
We attribute this to the knowledge sensitivity inherent in commonsense reasoning, 
where step-by-step reasoning cannot compensate for the lack of knowledge.
In event temporal reasoning, improvements mainly stem from datasets involving implicit multi-step reasoning (MenatQA and TempReason), indicating that CoT is more effective for multi-hop questions.
In summary, zero-shot CoT consistently has a negative impact on temporal reasoning.
While in few-shot scenario, CoT has a positive impact on symbolic and complex tasks, while negatively affecting knowledge-sensitive tasks.

\section{Analysis and Discussion}
\label{sec:analysis}

\subsection{Scaling Effect of Model Size}
\begin{figure}[t]
    \centering
    \includegraphics[clip, width=\linewidth]{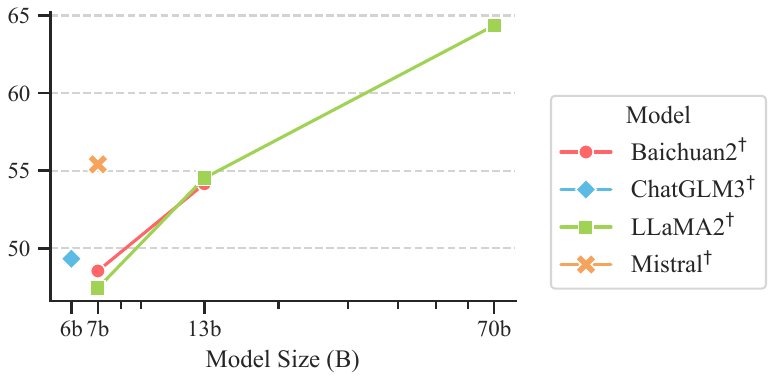}
    \caption{\label{fig:pareto}
    Scaling effect of model size and overall temporal reasoning performance. The x-axis (model size) is shown in the log scale. Results show a log-linearity between parameter size and performance.}
\end{figure}

We investigate how the scale of models affects temporal reasoning capabilities.
The trend is illustrated in Figure~\ref{fig:pareto}.
As the model scale increases, there is a notable improvement in performance. 
When the parameter size expands from 7B to 13B, LLaMA2 and Baichuan2 show improvements of 13.0\% and 10.5\%, respectively.
Furthermore, when LLaMA scales up to 70B, the trend of performance improvement continues without stopping.
The overall improvement follows a log-linear relationship with scale.
There are no significant performance differences among LLaMA2, Baichuan2, and ChatGLM3 under similar parameter specifications, 
while Mistral demonstrates impressive prowess, outperforming all other 13B models with nearly half the number of parameters.

\subsection{Challenges in Temporal Reasoning}
\definecolor{darkpastelgreen}{rgb}{0.01, 0.75, 0.24}
\newcommand{\floatingtext}[1]{\kern-0.1em\makebox[0pt][l]{#1}}
\newcommand{\increase}[1]{#1\floatingtext{\textcolor{darkpastelgreen}{\small~$\uparrow$}}}
\newcommand{\decrease}[1]{#1\floatingtext{\textcolor{red}{\small~$\downarrow$}}}

\begin{table}[t]
\centering
\resizebox{\linewidth}{!}{
\begin{tabular}{l|ccccc|c}
\toprule
Model & Order & Duration & Freq. & Stationarity & Typical & Avg. \\
\midrule
GPT-4 & \decrease{76.4}& \increase{92.8}& \increase{83.3}& \decrease{71.4}& \decrease{54.5}& 77.5 \\
GPT-3.5 & \increase{50.5}& \decrease{39.8}& \increase{55.2}& \increase{48.4}& \decrease{28.7}& 43.5 \\
Baichuan2$^\dag_{\text{13}b}$ & \decrease{40.5}& \increase{51.8}& \increase{43.7}& \increase{46.2}& \decrease{29.8}& 42.5 \\
LLaMA2$^\dag_{\text{70}b}$ 	& \increase{65.2} & \increase{72.1} & \increase{66.3} & \decrease{36.3} & \decrease{52.7} & 63.0 \\
Mistral$^\dag_{\text{7}b}$ & \decrease{27.0}& \increase{44.4}& \increase{58.3}& \decrease{38.5}& \decrease{38.3}& 42.5 \\
\bottomrule
\end{tabular}
}
\caption{\label{tab:analysis-commonsense}
Results in each temporal commonsense aspect under few-shot setting. Models with \dag~ are base models. Red\textcolor{red}{\small~$\downarrow$} and Green\textcolor{darkpastelgreen}{\small~$\uparrow$} represent the performance is lower or higher than its average performance.
Metric is EM.
}
\end{table}

\paragraph{LLMs underperform in (multi-hop) symbolic reasoning}
Except for GPT-4, the performance of all other models in symbolic temporal reasoning is unsatisfactory.
A noticeable decrease is observed in duration-conversion task compared to other atomic tasks (25\% in GPT-4 and 27\% in LLaMA2$_{\text{70}b}$).
This is because the duration-conversion task (s3) necessitates a two-step reasoning process.
It first unifies time units, and subsequently engages in numerical comparison.
In contrast, other atomic tasks (s1, s2 and arithmetic) can be completed with a single reasoning step.
In summary, LLMs perform poorly in symbolic temporal reasoning and exhibit more pronounced declines when encountering multi-step reasoning.

\paragraph{Mastery of commonsense knowledge varies in LLMs}
We analyze models' performance across various commonsense aspects, as shown in Table~\ref{tab:analysis-commonsense}.
We regard the model's average performance in commonsense reasoning tasks as the baseline.
If the model outperforms the baseline in a specific aspect, it suggests greater proficiency in this type of knowledge, and vice versa.
The findings indicate that LLMs generally demonstrate good knowledge of event duration and frequency. 
However, their comprehension of event order and typical events is relatively weaker
The uneven mastery of commonsense knowledge significantly affects the model's reasoning performance, 
especially when dealing with complex questions that involve multiple types of knowledge.
Retrieval-augmented reasoning presents a promising avenue for mitigating the model's knowledge scarcity.

\paragraph{LLMs exhibit poor implicit temporal reasoning capabilities.}
When comparing explicit and implicit event temporal reasoning, specifically TimeQA-explicit versus others, we observe a significant performance decrease in implicit reasoning.
Additionally, on TRACIE with numerous implied events, most models only surpass a random baseline (50.0). 
Even GPT-4, despite its advanced capabilities, achieves only a 66.4\% accuracy, suggesting that the LLM struggles with modeling implicit temporal relationships.
We consider it helpful to explicitly model the temporal relationships between events and time expressions, for instance constructing timelines or temporal graphs.

\paragraph{LLMs are good factual reasoners rather than factual extractors}
When humans engage in temporal reasoning, it generally involves two steps: first, extracting time-fact pairs from the context, and then performing fact-based reasoning.
TempReason provides extracted facts for conducting fact-based reasoning.
By comparing the model's performance in context-based (TimeQA) against fact-based (TempReason) reasoning, 
we identify the bottleneck in event temporal reasoning.
LLMs excel in TempReason, which signifies their strong capability in fact-based reasoning.
However, their performance in context-based reasoning is significantly weaker compared to their performance in fact-based reasoning.
This implies that errors could arise during the extraction of time-sensitive facts from the context.
We attribute this performance gap to the model's deficiency in factual extraction capabilities
Thus, we consider LLMs to be strong factual reasoners rather than factual extractors in event temporal reasoning.

\begin{figure}[t]
    \centering
    \includegraphics[clip, width=\linewidth]{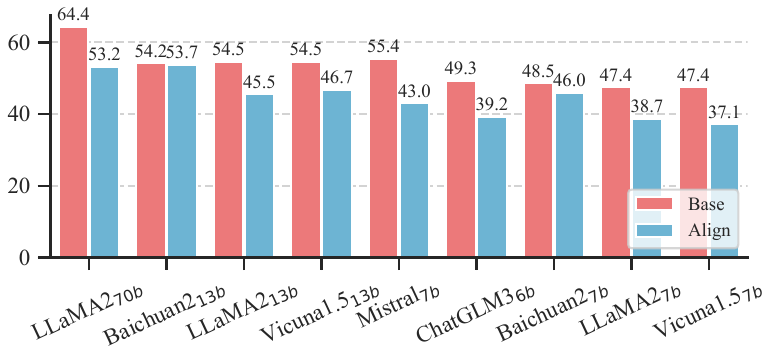}
    \caption{\label{figure:base-align}
    Performance difference between base and alignment models under few-shot setting.
    Baichuan2 and LLaMA2 are aligned with SFT and RLHF. Vicuna, Mistral and ChatGLM3 are aligned with only SFT.}
\end{figure}

\subsection{Alignment Impairs Temporal Reasoning}
In the experiments mentioned earlier (Table~\ref{tab:zeroshot_results}), we observe a sharp decline in zero-shot performance of alignment models.
To investigate whether alignment is the cause of the decline in temporal reasoning, we conducted experiments on alignment models under few-shot settings.
Figure~\ref{figure:base-align} illustrates the overall performance decline after alignment.
With the exception of Baichuan2, all other models are severely impaired, experiencing a significant drop of up to 22\%.
Through manual analysis of error cases, we have summarized two reasons:
(1) Alignment reduces the model's usability, causing it to tend towards refusal to answer when confronted with knowledge-sensitive questions.
(2) Alignment damages the model's in-context learning capability, resulting in situations where the model deviates from the demonstrations.
Furthermore, we believe that the lack of temporal reasoning-related training data in alignment exacerbates this issue, leading to disparities between different reasoning capabilities, such as mathematical and temporal reasoning.

\begin{figure*}[t]
    \centering
    \includegraphics[clip, width=\linewidth]{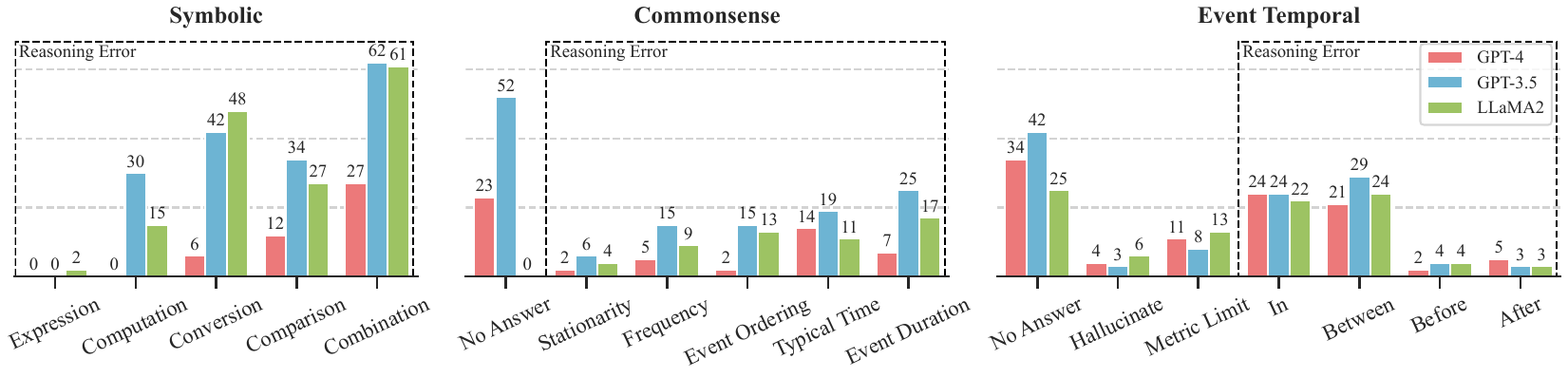}
    \caption{\label{figure:error-analyze}
    Error analysis for Symbolic, Commonsense, and Event Temporal. We select 100 test samples from each subtask for GPT-4, GPT-3.5 and LLaMa2-base$_{70b}$.}
\end{figure*}
\subsection{Error Analysis}
\label{appendix:error_analysis}
We manually analyze 100 predictions by GPT-4, GPT-3.5 and LLaMa2-base$_{70b}$ from each subtask.
The visualization of errors is shown in Figure ~\ref{figure:error-analyze}.

\paragraph{Symbolic Reasoning}
We categorize symbolic reasoning errors into five groups:
(a) \textit{Expression}: The model provides an incorrect time calculation expression.
(b) \textit{Computation}: The model provides the correct time calculation expression, but there is a calculation error.
(c) \textit{Conversion}: The model has an error in the conversion of time units.
(d) \textit{Comparison}: The model has an error when comparing two time-expressions (or intervals).
(e) \textit{Combination}: The model encounters errors in the combination of multiple above operations.
LLMs exhibit numerous computation, conversion, and comparison errors, which suggests a substantial deficiency in their understanding of fundamental temporal expressions.
Additionally, a higher frequency of errors is observed in combination questions, highlighting that multi-step reasoning continues to be a significant challenge for current models

\paragraph{Commonsense Reasoning}
We categorize the errors of commonsense reasoning into two groups: 
(a) \textit{No Answer}: The model fails to provide a final answer.
(b) \textit{Reasoning Error}: The model encounters reasoning errors, which can be subdivided into five types of knowledge-related errors.
We observe that GPT series models have a higher \textit{No Answer} rate, while LLaMA is always able to provide answers.
This discrepancy can be attributed to two factors: 
firstly, the models may lack the necessary commonsense knowledge to formulate an answer; 
secondly, the preference alignment mechanism may prompt the model to abstain from answering when confronted with questions outside its knowledge scope.
Integration of retrieval can alleviate the problem of knowledge scarcity to a certain degree.

\paragraph{Event Temporal Reasoning}
We categorize the errors of event temporal reasoning into four groups: 
(a) \textit{No Answer}: The model is unable to find the answer in the context.
(b) \textit{Reasoning Error}: The model encounters reasoning errors.
(c) \textit{Hallucination}: The model's prediction does not exist in the context, known as hallucination reasoning.
(d) \textit{Metric}: The model's prediction is correct, but the metric is limited by the evaluation criteria.
It can be observed that, except for reasoning errors, failures to provide answers account for approximately 30\%, 
indicating that models still have flaws in grounding temporal facts from context.
Additionally, models occasionally experience hallucination phenomena, leading to erroneous reasoning.

\section{Related Work \done}
\label{sec:relatedwork}

\subsection{Temporal Reasoning}
There are numerous efforts addressing diverse challenges in temporal reasoning.
Early research mainly relies on TimeML~\citep{timeml}, focusing TimeX extraction and temporal relation extraction~\citep{tempeval-1,tempeval-2,tempeval-3,tempevalqa, timex-extraction-1, temporal-relation-extraction-1,temporal-relation-extraction-2}. 
The advent of pre-trained language models (PLMs) has brought about commonsense reasoning as a tool to explore the world knowledge in models~\citep{mctaco,timedial,templama}.
Recently, much attention has shifted towards event temporal reasoning~\citep{timeqa, tempreason,menatqa}.
\citet{tft-econet,tft-once-opon-time,tft-replearn,tft-tml} continuously pre-trains LLMs on time-aware data to elicit temporal reasoning, 
and \citet{emt-program,emt-tempgraph-su,emt-tempgraph-chu} explicitly represent temporal relationships using temporal graphs and timelines.
Additionally, some works extend beyond text, evaluating temporal reasoning in structured tables and video domains~\citep{table-tqa, video-tqa}.

Some concurrent studies also analyze the temporal reasoning abilities of LLMs.
\citet{analysis-temp-commonsense-1,analysis-temp-commonsense-2} focus on temporal commonsense and \citet{tram} introduces a unified form for accessing the overall abilities.

Distinguished from other works, \ours{} is multispectral, offering a comprehensive evaluation of LLM's temporal reasoning abilities.

\subsection{Large Language Models}
In recent years, there has been rapid progress in the research of large language models (LLM)~\citep{llmsurvey}.
They exhibit outstanding performance across a multitude of tasks without the need for fine-tuning~\citep{gpt3,zeroshotcot}.
Furthermore, they have achieved astonishing results in complex reasoning tasks, such as mathematical reasoning~\citep{gsm8k,lila} and logical reasoning~\citep{reclor,glore}. 
Moreover, some studies suggest that the chain-of-thought prompting can further enhance the model's capabilities in complex reasoning scenarios~\citep{fewshotcot,zeroshotcot,cotsurvey,ignitingCot}.

\section{Conclusion}
\label{sec:conclusion}
Temporal reasoning entails inherent diversity and complexity.
The lack of a comprehensive benchmark makes it challenging to quantify LLMs' temporal reasoning capabilities.
In this work, we present \ours{}, a comprehensive and hierarchical benchmark for LLM temporal reasoning, tailored to mirror temporal reasoning in complex scenarios.
We conduct extensive experiments on state-of-the-art LLMs to investigate their temporal reasoning capabilities.
Our findings indicate a substantial gap between state-of-the-art LLMs and human performance, emphasizing the need for further research in this area.
Moreover, we provide a meticulous analysis and discussion, outlining the current challenges that models face and suggesting potential directions for improvement.
\section*{Limitations}
\label{sec:limitations}
TimeBench is a comprehensive benchmark to quantify the temporal reasoning capabilities of LLMs.
While we have taken various factors into account, there are a few limitations.
Firstly, our evaluation only applied prompt-based method under zero-shot and few-shot setting, lacking evaluations specifically tailored for models fine-tuned on the temporal domain.
Secondly, the instructions and demonstrations were manually crafted, 
which may potentially lead to discrepancies in prompts interpretation among different LLMs.
Thirdly, the dataset constituting the benchmark includes data from past years and a portion sourced from Wikipedia, which may contaminate the training corpus of LLMs.

\section*{Acknowledgements}
\label{sec:acknowledgements}
The research in this article is supported by the National Key Research and Development Project (2021YFF0901602), the National Science Foundation of China (U22B2059, 62276083), and Shenzhen Foundational Research Funding (JCYJ20200109113441941), Major Key Project of PCL (PCL2021A06).
Ming Liu is the corresponding author.

\bibliography{anthology,custom}

\begin{thebibliography}{59}
\providecommand{\natexlab}[1]{#1}

\bibitem[{Ainslie et~al.(2023)Ainslie, Lee{-}Thorp, de~Jong, Zemlyanskiy, Lebr{\'{o}}n, and Sanghai}]{gqa}
Joshua Ainslie, James Lee{-}Thorp, Michiel de~Jong, Yury Zemlyanskiy, Federico Lebr{\'{o}}n, and Sumit Sanghai. 2023.
\newblock \href {https://doi.org/10.18653/V1/2023.EMNLP-MAIN.298} {{GQA:} training generalized multi-query transformer models from multi-head checkpoints}.
\newblock In \emph{Proceedings of the 2023 Conference on Empirical Methods in Natural Language Processing, {EMNLP} 2023, Singapore, December 6-10, 2023}, pages 4895--4901. Association for Computational Linguistics.

\bibitem[{Banerjee and Lavie(2005)}]{meteor}
Satanjeev Banerjee and Alon Lavie. 2005.
\newblock Meteor: An automatic metric for mt evaluation with improved correlation with human judgments.
\newblock In \emph{Proceedings of the acl workshop on intrinsic and extrinsic evaluation measures for machine translation and/or summarization}, pages 65--72.

\bibitem[{Barsalou et~al.(2018)Barsalou, Dutriaux, and Scheepers}]{psychology}
Lawrence~W Barsalou, L{\'e}o Dutriaux, and Christoph Scheepers. 2018.
\newblock Moving beyond the distinction between concrete and abstract concepts.
\newblock \emph{Philosophical Transactions of the Royal Society B: Biological Sciences}, 373(1752):20170144.

\bibitem[{Beltagy et~al.(2020)Beltagy, Peters, and Cohan}]{swa}
Iz~Beltagy, Matthew~E Peters, and Arman Cohan. 2020.
\newblock Longformer: The long-document transformer.
\newblock \emph{arXiv preprint arXiv:2004.05150}.

\bibitem[{Brown et~al.(2020)Brown, Mann, Ryder, Subbiah, Kaplan, Dhariwal, Neelakantan, Shyam, Sastry, Askell, Agarwal, Herbert{-}Voss, Krueger, Henighan, Child, Ramesh, Ziegler, Wu, Winter, Hesse, Chen, Sigler, Litwin, Gray, Chess, Clark, Berner, McCandlish, Radford, Sutskever, and Amodei}]{gpt3}
Tom~B. Brown, Benjamin Mann, Nick Ryder, Melanie Subbiah, Jared Kaplan, Prafulla Dhariwal, Arvind Neelakantan, Pranav Shyam, Girish Sastry, Amanda Askell, Sandhini Agarwal, Ariel Herbert{-}Voss, Gretchen Krueger, Tom Henighan, Rewon Child, Aditya Ramesh, Daniel~M. Ziegler, Jeffrey Wu, Clemens Winter, Christopher Hesse, Mark Chen, Eric Sigler, Mateusz Litwin, Scott Gray, Benjamin Chess, Jack Clark, Christopher Berner, Sam McCandlish, Alec Radford, Ilya Sutskever, and Dario Amodei. 2020.
\newblock \href {https://proceedings.neurips.cc/paper/2020/hash/1457c0d6bfcb4967418bfb8ac142f64a-Abstract.html} {Language models are few-shot learners}.
\newblock In \emph{Advances in Neural Information Processing Systems 33: Annual Conference on Neural Information Processing Systems 2020, NeurIPS 2020, December 6-12, 2020, virtual}.

\bibitem[{Chen et~al.(2021)Chen, Wang, and Wang}]{timeqa}
Wenhu Chen, Xinyi Wang, and William~Yang Wang. 2021.
\newblock \href {https://datasets-benchmarks-proceedings.neurips.cc/paper/2021/hash/1f0e3dad99908345f7439f8ffabdffc4-Abstract-round2.html} {A dataset for answering time-sensitive questions}.
\newblock In \emph{Proceedings of the Neural Information Processing Systems Track on Datasets and Benchmarks 1, NeurIPS Datasets and Benchmarks 2021, December 2021, virtual}.

\bibitem[{Chen et~al.(2023)Chen, Wu, Zhang, and Feng}]{tft-tml}
Ziqiang Chen, Shaojuan Wu, Xiaowang Zhang, and Zhiyong Feng. 2023.
\newblock \href {https://doi.org/10.1145/3543873.3587347} {{TML:} {A} temporal-aware multitask learning framework for time-sensitive question answering}.
\newblock In \emph{Companion Proceedings of the {ACM} Web Conference 2023, {WWW} 2023, Austin, TX, USA, 30 April 2023 - 4 May 2023}, pages 200--203. {ACM}.

\bibitem[{Chiang et~al.(2023)Chiang, Li, Lin, Sheng, Wu, Zhang, Zheng, Zhuang, Zhuang, Gonzalez, Stoica, and Xing}]{vicuna:chiang}
Wei-Lin Chiang, Zhuohan Li, Zi~Lin, Ying Sheng, Zhanghao Wu, Hao Zhang, Lianmin Zheng, Siyuan Zhuang, Yonghao Zhuang, Joseph~E. Gonzalez, Ion Stoica, and Eric~P. Xing. 2023.
\newblock \href {https://lmsys.org/blog/2023-03-30-vicuna/} {Vicuna: An open-source chatbot impressing gpt-4 with 90\%* chatgpt quality}.

\bibitem[{Chowdhery et~al.(2023)Chowdhery, Narang, Devlin, Bosma, Mishra, Roberts, Barham, Chung, Sutton, Gehrmann, Schuh, Shi, Tsvyashchenko, Maynez, Rao, Barnes, Tay, Shazeer, Prabhakaran, Reif, Du, Hutchinson, Pope, Bradbury, Austin, Isard, Gur{-}Ari, Yin, Duke, Levskaya, Ghemawat, Dev, Michalewski, Garcia, Misra, Robinson, Fedus, Zhou, Ippolito, Luan, Lim, Zoph, Spiridonov, Sepassi, Dohan, Agrawal, Omernick, Dai, Pillai, Pellat, Lewkowycz, Moreira, Child, Polozov, Lee, Zhou, Wang, Saeta, Diaz, Firat, Catasta, Wei, Meier{-}Hellstern, Eck, Dean, Petrov, and Fiedel}]{palm}
Aakanksha Chowdhery, Sharan Narang, Jacob Devlin, Maarten Bosma, Gaurav Mishra, Adam Roberts, Paul Barham, Hyung~Won Chung, Charles Sutton, Sebastian Gehrmann, Parker Schuh, Kensen Shi, Sasha Tsvyashchenko, Joshua Maynez, Abhishek Rao, Parker Barnes, Yi~Tay, Noam Shazeer, Vinodkumar Prabhakaran, Emily Reif, Nan Du, Ben Hutchinson, Reiner Pope, James Bradbury, Jacob Austin, Michael Isard, Guy Gur{-}Ari, Pengcheng Yin, Toju Duke, Anselm Levskaya, Sanjay Ghemawat, Sunipa Dev, Henryk Michalewski, Xavier Garcia, Vedant Misra, Kevin Robinson, Liam Fedus, Denny Zhou, Daphne Ippolito, David Luan, Hyeontaek Lim, Barret Zoph, Alexander Spiridonov, Ryan Sepassi, David Dohan, Shivani Agrawal, Mark Omernick, Andrew~M. Dai, Thanumalayan~Sankaranarayana Pillai, Marie Pellat, Aitor Lewkowycz, Erica Moreira, Rewon Child, Oleksandr Polozov, Katherine Lee, Zongwei Zhou, Xuezhi Wang, Brennan Saeta, Mark Diaz, Orhan Firat, Michele Catasta, Jason Wei, Kathy Meier{-}Hellstern, Douglas Eck, Jeff Dean, Slav Petrov, and Noah Fiedel.
  2023.
\newblock \href {http://jmlr.org/papers/v24/22-1144.html} {Palm: Scaling language modeling with pathways}.
\newblock \emph{J. Mach. Learn. Res.}, 24:240:1--240:113.

\bibitem[{Chu et~al.(2024)Chu, Chen, Chen, Yu, He, Wang, Peng, Liu, Qin, and Liu}]{cotsurvey}
Zheng Chu, Jingchang Chen, Qianglong Chen, Weijiang Yu, Tao He, Haotian Wang, Weihua Peng, Ming Liu, Bing Qin, and Ting Liu. 2024.
\newblock \href {https://arxiv.org/abs/2309.15402} {Navigate through enigmatic labyrinth a survey of chain of thought reasoning: Advances, frontiers and future}.
\newblock In \emph{The 62nd Annual Meeting of the Association for Computational Linguistics: ACL 2024, Bangkok, Thailand, August 11–16, 2024}. Association for Computational Linguistics.

\bibitem[{Chu et~al.(2023)Chu, Wang, Liang, Liu, and Qin}]{emt-tempgraph-chu}
Zheng Chu, Zekun Wang, Jiafeng Liang, Ming Liu, and Bing Qin. 2023.
\newblock \href {https://doi.org/10.18653/V1/2023.FINDINGS-EMNLP.1016} {{MTGER:} multi-view temporal graph enhanced temporal reasoning over time-involved document}.
\newblock In \emph{Findings of the Association for Computational Linguistics: {EMNLP} 2023, Singapore, December 6-10, 2023}, pages 15218--15233. Association for Computational Linguistics.

\bibitem[{Chung et~al.(2022)Chung, Hou, Longpre, Zoph, Tay, Fedus, Li, Wang, Dehghani, Brahma, Webson, Gu, Dai, Suzgun, Chen, Chowdhery, Castro-Ros, Pellat, Robinson, Valter, Narang, Mishra, Yu, Zhao, Huang, Dai, Yu, Petrov, Chi, Dean, Devlin, Roberts, Zhou, Le, and Wei}]{flant5}
Hyung~Won Chung, Le~Hou, Shayne Longpre, Barret Zoph, Yi~Tay, William Fedus, Yunxuan Li, Xuezhi Wang, Mostafa Dehghani, Siddhartha Brahma, Albert Webson, Shixiang~Shane Gu, Zhuyun Dai, Mirac Suzgun, Xinyun Chen, Aakanksha Chowdhery, Alex Castro-Ros, Marie Pellat, Kevin Robinson, Dasha Valter, Sharan Narang, Gaurav Mishra, Adams Yu, Vincent Zhao, Yanping Huang, Andrew Dai, Hongkun Yu, Slav Petrov, Ed~H. Chi, Jeff Dean, Jacob Devlin, Adam Roberts, Denny Zhou, Quoc~V. Le, and Jason Wei. 2022.
\newblock \href {https://arxiv.org/abs/2210.11416} {Scaling instruction-finetuned language models}.
\newblock \emph{Preprint}, arXiv:2210.11416.

\bibitem[{Cobbe et~al.(2021)Cobbe, Kosaraju, Bavarian, Chen, Jun, Kaiser, Plappert, Tworek, Hilton, Nakano, Hesse, and Schulman}]{gsm8k}
Karl Cobbe, Vineet Kosaraju, Mohammad Bavarian, Mark Chen, Heewoo Jun, Lukasz Kaiser, Matthias Plappert, Jerry Tworek, Jacob Hilton, Reiichiro Nakano, Christopher Hesse, and John Schulman. 2021.
\newblock \href {https://arxiv.org/abs/2110.14168} {Training verifiers to solve math word problems}.
\newblock \emph{CoRR}, abs/2110.14168.

\bibitem[{Dhingra et~al.(2022)Dhingra, Cole, Eisenschlos, Gillick, Eisenstein, and Cohen}]{templama}
Bhuwan Dhingra, Jeremy~R. Cole, Julian~Martin Eisenschlos, Daniel Gillick, Jacob Eisenstein, and William~W. Cohen. 2022.
\newblock \href {https://doi.org/10.1162/TACL\_A\_00459} {Time-aware language models as temporal knowledge bases}.
\newblock \emph{Trans. Assoc. Comput. Linguistics}, 10:257--273.

\bibitem[{Gupta et~al.(2023)Gupta, Kandoi, Vora, Zhang, He, Reinanda, and Srikumar}]{table-tqa}
Vivek Gupta, Pranshu Kandoi, Mahek~Bhavesh Vora, Shuo Zhang, Yujie He, Ridho Reinanda, and Vivek Srikumar. 2023.
\newblock \href {https://doi.org/10.18653/V1/2023.EMNLP-MAIN.149} {Temptabqa: Temporal question answering for semi-structured tables}.
\newblock In \emph{Proceedings of the 2023 Conference on Empirical Methods in Natural Language Processing, {EMNLP} 2023, Singapore, December 6-10, 2023}, pages 2431--2453. Association for Computational Linguistics.

\bibitem[{Han et~al.(2021)Han, Ren, and Peng}]{tft-econet}
Rujun Han, Xiang Ren, and Nanyun Peng. 2021.
\newblock \href {https://doi.org/10.18653/V1/2021.EMNLP-MAIN.436} {{ECONET:} effective continual pretraining of language models for event temporal reasoning}.
\newblock In \emph{Proceedings of the 2021 Conference on Empirical Methods in Natural Language Processing, {EMNLP} 2021, Virtual Event / Punta Cana, Dominican Republic, 7-11 November, 2021}, pages 5367--5380. Association for Computational Linguistics.

\bibitem[{Hendrycks et~al.(2021)Hendrycks, Burns, Basart, Zou, Mazeika, Song, and Steinhardt}]{mmlu}
Dan Hendrycks, Collin Burns, Steven Basart, Andy Zou, Mantas Mazeika, Dawn Song, and Jacob Steinhardt. 2021.
\newblock \href {https://openreview.net/forum?id=d7KBjmI3GmQ} {Measuring massive multitask language understanding}.
\newblock In \emph{9th International Conference on Learning Representations, {ICLR} 2021, Virtual Event, Austria, May 3-7, 2021}. OpenReview.net.

\bibitem[{Jain et~al.(2023)Jain, Sojitra, Acharya, Saha, Jatowt, and Dandapat}]{analysis-temp-commonsense-1}
Raghav Jain, Daivik Sojitra, Arkadeep Acharya, Sriparna Saha, Adam Jatowt, and Sandipan Dandapat. 2023.
\newblock \href {https://doi.org/10.18653/v1/2023.emnlp-main.418} {Do language models have a common sense regarding time? revisiting temporal commonsense reasoning in the era of large language models}.
\newblock In \emph{Proceedings of the 2023 Conference on Empirical Methods in Natural Language Processing}, pages 6750--6774, Singapore. Association for Computational Linguistics.

\bibitem[{Jiang et~al.(2023)Jiang, Sablayrolles, Mensch, Bamford, Chaplot, de~Las~Casas, Bressand, Lengyel, Lample, Saulnier, Lavaud, Lachaux, Stock, Scao, Lavril, Wang, Lacroix, and Sayed}]{mistral}
Albert~Q. Jiang, Alexandre Sablayrolles, Arthur Mensch, Chris Bamford, Devendra~Singh Chaplot, Diego de~Las~Casas, Florian Bressand, Gianna Lengyel, Guillaume Lample, Lucile Saulnier, L{\'{e}}lio~Renard Lavaud, Marie{-}Anne Lachaux, Pierre Stock, Teven~Le Scao, Thibaut Lavril, Thomas Wang, Timoth{\'{e}}e Lacroix, and William~El Sayed. 2023.
\newblock \href {https://doi.org/10.48550/ARXIV.2310.06825} {Mistral 7b}.
\newblock \emph{CoRR}, abs/2310.06825.

\bibitem[{Ko et~al.(2023)Ko, Lee, Kang, Roh, and Kim}]{video-tqa}
Dohwan Ko, Ji~Soo Lee, Woo{-}Young Kang, Byungseok Roh, and Hyunwoo Kim. 2023.
\newblock \href {https://doi.org/10.18653/V1/2023.EMNLP-MAIN.261} {Large language models are temporal and causal reasoners for video question answering}.
\newblock In \emph{Proceedings of the 2023 Conference on Empirical Methods in Natural Language Processing, {EMNLP} 2023, Singapore, December 6-10, 2023}, pages 4300--4316. Association for Computational Linguistics.

\bibitem[{Kojima et~al.(2022)Kojima, Gu, Reid, Matsuo, and Iwasawa}]{zeroshotcot}
Takeshi Kojima, Shixiang~Shane Gu, Machel Reid, Yutaka Matsuo, and Yusuke Iwasawa. 2022.
\newblock \href {http://papers.nips.cc/paper\_files/paper/2022/hash/8bb0d291acd4acf06ef112099c16f326-Abstract-Conference.html} {Large language models are zero-shot reasoners}.
\newblock In \emph{NeurIPS}.

\bibitem[{Lin(2004)}]{rouge}
Chin-Yew Lin. 2004.
\newblock Rouge: A package for automatic evaluation of summaries.
\newblock In \emph{Text summarization branches out}, pages 74--81.

\bibitem[{Liu et~al.(2023)Liu, Teng, Ning, Liu, Zhou, and Zhang}]{glore}
Hanmeng Liu, Zhiyang Teng, Ruoxi Ning, Jian Liu, Qiji Zhou, and Yue Zhang. 2023.
\newblock \href {https://doi.org/10.48550/ARXIV.2310.09107} {Glore: Evaluating logical reasoning of large language models}.
\newblock \emph{CoRR}, abs/2310.09107.

\bibitem[{Llorens et~al.(2015)Llorens, Chambers, UzZaman, Mostafazadeh, Allen, and Pustejovsky}]{tempevalqa}
Hector Llorens, Nathanael Chambers, Naushad UzZaman, Nasrin Mostafazadeh, James~F. Allen, and James Pustejovsky. 2015.
\newblock \href {https://doi.org/10.18653/V1/S15-2134} {Semeval-2015 task 5: {QA} tempeval - evaluating temporal information understanding with question answering}.
\newblock In \emph{Proceedings of the 9th International Workshop on Semantic Evaluation, SemEval@NAACL-HLT 2015, Denver, Colorado, USA, June 4-5, 2015}, pages 792--800. The Association for Computer Linguistics.

\bibitem[{Mathur et~al.(2021)Mathur, Jain, Dernoncourt, Morariu, Tran, and Manocha}]{temporal-relation-extraction-1}
Puneet Mathur, Rajiv Jain, Franck Dernoncourt, Vlad Morariu, Quan~Hung Tran, and Dinesh Manocha. 2021.
\newblock \href {https://doi.org/10.18653/v1/2021.acl-short.67} {{TIMERS}: Document-level temporal relation extraction}.
\newblock In \emph{Proceedings of the 59th Annual Meeting of the Association for Computational Linguistics and the 11th International Joint Conference on Natural Language Processing (Volume 2: Short Papers)}, pages 524--533, Online. Association for Computational Linguistics.

\bibitem[{Miller et~al.(2015)Miller, Bethard, Dligach, Lin, and Savova}]{timex-extraction-1}
Timothy Miller, Steven Bethard, Dmitriy Dligach, Chen Lin, and Guergana Savova. 2015.
\newblock \href {https://doi.org/10.18653/v1/W15-3809} {Extracting time expressions from clinical text}.
\newblock In \emph{Proceedings of {B}io{NLP} 15}, pages 81--91, Beijing, China. Association for Computational Linguistics.

\bibitem[{Mishra et~al.(2022)Mishra, Finlayson, Lu, Tang, Welleck, Baral, Rajpurohit, Tafjord, Sabharwal, Clark, and Kalyan}]{lila}
Swaroop Mishra, Matthew Finlayson, Pan Lu, Leonard Tang, Sean Welleck, Chitta Baral, Tanmay Rajpurohit, Oyvind Tafjord, Ashish Sabharwal, Peter Clark, and Ashwin Kalyan. 2022.
\newblock \href {https://doi.org/10.18653/V1/2022.EMNLP-MAIN.392} {{LILA:} {A} unified benchmark for mathematical reasoning}.
\newblock In \emph{Proceedings of the 2022 Conference on Empirical Methods in Natural Language Processing, {EMNLP} 2022, Abu Dhabi, United Arab Emirates, December 7-11, 2022}, pages 5807--5832. Association for Computational Linguistics.

\bibitem[{OpenAI(2023)}]{gpt4}
OpenAI. 2023.
\newblock \href {https://doi.org/10.48550/ARXIV.2303.08774} {{GPT-4} technical report}.
\newblock \emph{CoRR}, abs/2303.08774.

\bibitem[{Ouyang et~al.(2022)Ouyang, Wu, Jiang, Almeida, Wainwright, Mishkin, Zhang, Agarwal, Slama, Ray, Schulman, Hilton, Kelton, Miller, Simens, Askell, Welinder, Christiano, Leike, and Lowe}]{gpt3.5}
Long Ouyang, Jeffrey Wu, Xu~Jiang, Diogo Almeida, Carroll~L. Wainwright, Pamela Mishkin, Chong Zhang, Sandhini Agarwal, Katarina Slama, Alex Ray, John Schulman, Jacob Hilton, Fraser Kelton, Luke Miller, Maddie Simens, Amanda Askell, Peter Welinder, Paul~F. Christiano, Jan Leike, and Ryan Lowe. 2022.
\newblock \href {http://papers.nips.cc/paper\_files/paper/2022/hash/b1efde53be364a73914f58805a001731-Abstract-Conference.html} {Training language models to follow instructions with human feedback}.
\newblock In \emph{NeurIPS}.

\bibitem[{Papineni et~al.(2002)Papineni, Roukos, Ward, and Zhu}]{bleu}
Kishore Papineni, Salim Roukos, Todd Ward, and Wei-Jing Zhu. 2002.
\newblock Bleu: a method for automatic evaluation of machine translation.
\newblock In \emph{Proceedings of the 40th annual meeting of the Association for Computational Linguistics}, pages 311--318.

\bibitem[{Pustejovsky et~al.(2003)Pustejovsky, Casta{\~{n}}o, Ingria, Saur{\'{\i}}, Gaizauskas, Setzer, Katz, and Radev}]{timeml}
James Pustejovsky, Jos{\'{e}}~M. Casta{\~{n}}o, Robert Ingria, Roser Saur{\'{\i}}, Robert~J. Gaizauskas, Andrea Setzer, Graham Katz, and Dragomir~R. Radev. 2003.
\newblock Timeml: Robust specification of event and temporal expressions in text.
\newblock In \emph{New Directions in Question Answering, Papers from 2003 {AAAI} Spring Symposium, Stanford University, Stanford, CA, {USA}}, pages 28--34. {AAAI} Press.

\bibitem[{Qin et~al.(2021)Qin, Gupta, Upadhyay, He, Choi, and Faruqui}]{timedial}
Lianhui Qin, Aditya Gupta, Shyam Upadhyay, Luheng He, Yejin Choi, and Manaal Faruqui. 2021.
\newblock \href {https://doi.org/10.18653/V1/2021.ACL-LONG.549} {{TIMEDIAL:} temporal commonsense reasoning in dialog}.
\newblock In \emph{Proceedings of the 59th Annual Meeting of the Association for Computational Linguistics and the 11th International Joint Conference on Natural Language Processing, {ACL/IJCNLP} 2021, (Volume 1: Long Papers), Virtual Event, August 1-6, 2021}, pages 7066--7076. Association for Computational Linguistics.

\bibitem[{Qiu et~al.(2023)Qiu, Zhao, Ziser, Korhonen, Ponti, and Cohen}]{analysis-temp-commonsense-2}
Yifu Qiu, Zheng Zhao, Yftah Ziser, Anna Korhonen, Edoardo~M. Ponti, and Shay~B. Cohen. 2023.
\newblock \href {https://doi.org/10.48550/ARXIV.2311.08398} {Are large language models temporally grounded?}
\newblock \emph{CoRR}, abs/2311.08398.

\bibitem[{Raffel et~al.(2020)Raffel, Shazeer, Roberts, Lee, Narang, Matena, Zhou, Li, and Liu}]{t5}
Colin Raffel, Noam Shazeer, Adam Roberts, Katherine Lee, Sharan Narang, Michael Matena, Yanqi Zhou, Wei Li, and Peter~J. Liu. 2020.
\newblock \href {http://jmlr.org/papers/v21/20-074.html} {Exploring the limits of transfer learning with a unified text-to-text transformer}.
\newblock \emph{J. Mach. Learn. Res.}, 21:140:1--140:67.

\bibitem[{Son and Oh(2023)}]{tft-replearn}
Jungbin Son and Alice Oh. 2023.
\newblock \href {https://doi.org/10.18653/V1/2023.FINDINGS-EMNLP.6} {Time-aware representation learning for time-sensitive question answering}.
\newblock In \emph{Findings of the Association for Computational Linguistics: {EMNLP} 2023, Singapore, December 6-10, 2023}, pages 70--77. Association for Computational Linguistics.

\bibitem[{Srivastava et~al.(2022)Srivastava, Rastogi, Rao, Shoeb, Abid, Fisch, Brown, Santoro, Gupta, Garriga{-}Alonso, Kluska, Lewkowycz, Agarwal, Power, Ray, Warstadt, Kocurek, Safaya, Tazarv, Xiang, Parrish, Nie, Hussain, Askell, Dsouza, Rahane, Iyer, Andreassen, Santilli, Stuhlm{\"{u}}ller, Dai, La, Lampinen, Zou, Jiang, Chen, Vuong, Gupta, Gottardi, Norelli, Venkatesh, Gholamidavoodi, Tabassum, Menezes, Kirubarajan, Mullokandov, Sabharwal, Herrick, Efrat, Erdem, Karakas, and et~al.}]{bigbench}
Aarohi Srivastava, Abhinav Rastogi, Abhishek Rao, Abu Awal~Md Shoeb, Abubakar Abid, Adam Fisch, Adam~R. Brown, Adam Santoro, Aditya Gupta, Adri{\`{a}} Garriga{-}Alonso, Agnieszka Kluska, Aitor Lewkowycz, Akshat Agarwal, Alethea Power, Alex Ray, Alex Warstadt, Alexander~W. Kocurek, Ali Safaya, Ali Tazarv, Alice Xiang, Alicia Parrish, Allen Nie, Aman Hussain, Amanda Askell, Amanda Dsouza, Ameet Rahane, Anantharaman~S. Iyer, Anders Andreassen, Andrea Santilli, Andreas Stuhlm{\"{u}}ller, Andrew~M. Dai, Andrew La, Andrew~K. Lampinen, Andy Zou, Angela Jiang, Angelica Chen, Anh Vuong, Animesh Gupta, Anna Gottardi, Antonio Norelli, Anu Venkatesh, Arash Gholamidavoodi, Arfa Tabassum, Arul Menezes, Arun Kirubarajan, Asher Mullokandov, Ashish Sabharwal, Austin Herrick, Avia Efrat, Aykut Erdem, Ayla Karakas, and et~al. 2022.
\newblock \href {https://doi.org/10.48550/ARXIV.2206.04615} {Beyond the imitation game: Quantifying and extrapolating the capabilities of language models}.
\newblock \emph{CoRR}, abs/2206.04615.

\bibitem[{Su et~al.(2023)Su, Howard, Hakim, and Bethard}]{emt-tempgraph-su}
Xin Su, Phillip Howard, Nagib Hakim, and Steven Bethard. 2023.
\newblock \href {https://doi.org/10.18653/V1/2023.FINDINGS-EMNLP.67} {Fusing temporal graphs into transformers for time-sensitive question answering}.
\newblock In \emph{Findings of the Association for Computational Linguistics: {EMNLP} 2023, Singapore, December 6-10, 2023}, pages 948--966. Association for Computational Linguistics.

\bibitem[{Tan et~al.(2023)Tan, Ng, and Bing}]{tempreason}
Qingyu Tan, Hwee~Tou Ng, and Lidong Bing. 2023.
\newblock \href {https://doi.org/10.18653/V1/2023.ACL-LONG.828} {Towards benchmarking and improving the temporal reasoning capability of large language models}.
\newblock In \emph{Proceedings of the 61st Annual Meeting of the Association for Computational Linguistics (Volume 1: Long Papers), {ACL} 2023, Toronto, Canada, July 9-14, 2023}, pages 14820--14835. Association for Computational Linguistics.

\bibitem[{Thukral et~al.(2021)Thukral, Kukreja, and Kavouras}]{timexnli}
Shivin Thukral, Kunal Kukreja, and Christian Kavouras. 2021.
\newblock \href {https://doi.org/10.18653/V1/2021.BLACKBOXNLP-1.31} {Probing language models for understanding of temporal expressions}.
\newblock In \emph{Proceedings of the Fourth BlackboxNLP Workshop on Analyzing and Interpreting Neural Networks for NLP, BlackboxNLP@EMNLP 2021, Punta Cana, Dominican Republic, November 11, 2021}, pages 396--406. Association for Computational Linguistics.

\bibitem[{Touvron et~al.(2023)Touvron, Martin, Stone, Albert, Almahairi, Babaei, Bashlykov, Batra, Bhargava, Bhosale, Bikel, Blecher, Ferrer, Chen, Cucurull, Esiobu, Fernandes, Fu, Fu, Fuller, Gao, Goswami, Goyal, Hartshorn, Hosseini, Hou, Inan, Kardas, Kerkez, Khabsa, Kloumann, Korenev, Koura, Lachaux, Lavril, Lee, Liskovich, Lu, Mao, Martinet, Mihaylov, Mishra, Molybog, Nie, Poulton, Reizenstein, Rungta, Saladi, Schelten, Silva, Smith, Subramanian, Tan, Tang, Taylor, Williams, Kuan, Xu, Yan, Zarov, Zhang, Fan, Kambadur, Narang, Rodriguez, Stojnic, Edunov, and Scialom}]{llama2}
Hugo Touvron, Louis Martin, Kevin Stone, Peter Albert, Amjad Almahairi, Yasmine Babaei, Nikolay Bashlykov, Soumya Batra, Prajjwal Bhargava, Shruti Bhosale, Dan Bikel, Lukas Blecher, Cristian~Canton Ferrer, Moya Chen, Guillem Cucurull, David Esiobu, Jude Fernandes, Jeremy Fu, Wenyin Fu, Brian Fuller, Cynthia Gao, Vedanuj Goswami, Naman Goyal, Anthony Hartshorn, Saghar Hosseini, Rui Hou, Hakan Inan, Marcin Kardas, Viktor Kerkez, Madian Khabsa, Isabel Kloumann, Artem Korenev, Punit~Singh Koura, Marie-Anne Lachaux, Thibaut Lavril, Jenya Lee, Diana Liskovich, Yinghai Lu, Yuning Mao, Xavier Martinet, Todor Mihaylov, Pushkar Mishra, Igor Molybog, Yixin Nie, Andrew Poulton, Jeremy Reizenstein, Rashi Rungta, Kalyan Saladi, Alan Schelten, Ruan Silva, Eric~Michael Smith, Ranjan Subramanian, Xiaoqing~Ellen Tan, Binh Tang, Ross Taylor, Adina Williams, Jian~Xiang Kuan, Puxin Xu, Zheng Yan, Iliyan Zarov, Yuchen Zhang, Angela Fan, Melanie Kambadur, Sharan Narang, Aurelien Rodriguez, Robert Stojnic, Sergey Edunov, and Thomas
  Scialom. 2023.
\newblock \href {https://arxiv.org/abs/2307.09288} {Llama 2: Open foundation and fine-tuned chat models}.
\newblock \emph{Preprint}, arXiv:2307.09288.

\bibitem[{UzZaman et~al.(2013)UzZaman, Llorens, Derczynski, Allen, Verhagen, and Pustejovsky}]{tempeval-3}
Naushad UzZaman, Hector Llorens, Leon Derczynski, James~F. Allen, Marc Verhagen, and James Pustejovsky. 2013.
\newblock \href {https://aclanthology.org/S13-2001/} {Semeval-2013 task 1: Tempeval-3: Evaluating time expressions, events, and temporal relations}.
\newblock In \emph{Proceedings of the 7th International Workshop on Semantic Evaluation, SemEval@NAACL-HLT 2013, Atlanta, Georgia, USA, June 14-15, 2013}, pages 1--9. The Association for Computer Linguistics.

\bibitem[{Vashishtha et~al.(2019)Vashishtha, Van~Durme, and White}]{temporal-relation-extraction-2}
Siddharth Vashishtha, Benjamin Van~Durme, and Aaron~Steven White. 2019.
\newblock \href {https://doi.org/10.18653/v1/P19-1280} {Fine-grained temporal relation extraction}.
\newblock In \emph{Proceedings of the 57th Annual Meeting of the Association for Computational Linguistics}, pages 2906--2919, Florence, Italy. Association for Computational Linguistics.

\bibitem[{Vedantam et~al.(2015)Vedantam, Zitnick, and Parikh}]{cider}
Ramakrishna Vedantam, C.~Lawrence Zitnick, and Devi Parikh. 2015.
\newblock \href {https://doi.org/10.1109/CVPR.2015.7299087} {Cider: Consensus-based image description evaluation}.
\newblock In \emph{{IEEE} Conference on Computer Vision and Pattern Recognition, {CVPR} 2015, Boston, MA, USA, June 7-12, 2015}, pages 4566--4575. {IEEE} Computer Society.

\bibitem[{Verhagen et~al.(2007)Verhagen, Gaizauskas, Schilder, Hepple, Katz, and Pustejovsky}]{tempeval-1}
Marc Verhagen, Robert~J. Gaizauskas, Frank Schilder, Mark Hepple, Graham Katz, and James Pustejovsky. 2007.
\newblock \href {https://aclanthology.org/S07-1014/} {Semeval-2007 task 15: Tempeval temporal relation identification}.
\newblock In \emph{Proceedings of the 4th International Workshop on Semantic Evaluations, SemEval@ACL 2007, Prague, Czech Republic, June 23-24, 2007}, pages 75--80. The Association for Computer Linguistics.

\bibitem[{Verhagen et~al.(2010)Verhagen, Saur{\'{\i}}, Caselli, and Pustejovsky}]{tempeval-2}
Marc Verhagen, Roser Saur{\'{\i}}, Tommaso Caselli, and James Pustejovsky. 2010.
\newblock \href {https://aclanthology.org/S10-1010/} {Semeval-2010 task 13: Tempeval-2}.
\newblock In \emph{Proceedings of the 5th International Workshop on Semantic Evaluation, SemEval@ACL 2010, Uppsala University, Uppsala, Sweden, July 15-16, 2010}, pages 57--62. The Association for Computer Linguistics.

\bibitem[{Virgo et~al.(2022)Virgo, Cheng, and Kurohashi}]{durationqa}
Felix~Giovanni Virgo, Fei Cheng, and Sadao Kurohashi. 2022.
\newblock \href {https://aclanthology.org/2022.lrec-1.473} {Improving event duration question answering by leveraging existing temporal information extraction data}.
\newblock In \emph{Proceedings of the Thirteenth Language Resources and Evaluation Conference, {LREC} 2022, Marseille, France, 20-25 June 2022}, pages 4451--4457. European Language Resources Association.

\bibitem[{Wang and Zhao(2023)}]{tram}
Yuqing Wang and Yun Zhao. 2023.
\newblock \href {https://doi.org/10.48550/ARXIV.2310.00835} {{TRAM:} benchmarking temporal reasoning for large language models}.
\newblock \emph{CoRR}, abs/2310.00835.

\bibitem[{Wei et~al.(2022)Wei, Wang, Schuurmans, Bosma, Ichter, Xia, Chi, Le, and Zhou}]{fewshotcot}
Jason Wei, Xuezhi Wang, Dale Schuurmans, Maarten Bosma, Brian Ichter, Fei Xia, Ed~H. Chi, Quoc~V. Le, and Denny Zhou. 2022.
\newblock \href {http://papers.nips.cc/paper\_files/paper/2022/hash/9d5609613524ecf4f15af0f7b31abca4-Abstract-Conference.html} {Chain-of-thought prompting elicits reasoning in large language models}.
\newblock In \emph{NeurIPS}.

\bibitem[{Wei et~al.(2023)Wei, Su, Ma, Yu, Lei, Zhang, Zhao, and Liu}]{menatqa}
Yifan Wei, Yisong Su, Huanhuan Ma, Xiaoyan Yu, Fangyu Lei, Yuanzhe Zhang, Jun Zhao, and Kang Liu. 2023.
\newblock \href {https://doi.org/10.18653/V1/2023.FINDINGS-EMNLP.100} {Menatqa: {A} new dataset for testing the temporal comprehension and reasoning abilities of large language models}.
\newblock In \emph{Findings of the Association for Computational Linguistics: {EMNLP} 2023, Singapore, December 6-10, 2023}, pages 1434--1447. Association for Computational Linguistics.

\bibitem[{Yang et~al.(2023{\natexlab{a}})Yang, Xiao, Wang, Zhang, Bian, Yin, Lv, Pan, Wang, Yan, Yang, Deng, Wang, Liu, Ai, Dong, Zhao, Xu, Sun, Zhang, Liu, Ji, Xie, Dai, Fang, Su, Song, Liu, Ru, Ma, Wang, Liu, Lin, Nie, Guo, Sun, Zhang, Li, Li, Cheng, Chen, Zeng, Wang, Chen, Men, Yu, Pan, Shen, Wang, Li, Jiang, Gao, Zhang, Zhou, and Wu}]{baichuan2}
Aiyuan Yang, Bin Xiao, Bingning Wang, Borong Zhang, Ce~Bian, Chao Yin, Chenxu Lv, Da~Pan, Dian Wang, Dong Yan, Fan Yang, Fei Deng, Feng Wang, Feng Liu, Guangwei Ai, Guosheng Dong, Haizhou Zhao, Hang Xu, Haoze Sun, Hongda Zhang, Hui Liu, Jiaming Ji, Jian Xie, Juntao Dai, Kun Fang, Lei Su, Liang Song, Lifeng Liu, Liyun Ru, Luyao Ma, Mang Wang, Mickel Liu, MingAn Lin, Nuolan Nie, Peidong Guo, Ruiyang Sun, Tao Zhang, Tianpeng Li, Tianyu Li, Wei Cheng, Weipeng Chen, Xiangrong Zeng, Xiaochuan Wang, Xiaoxi Chen, Xin Men, Xin Yu, Xuehai Pan, Yanjun Shen, Yiding Wang, Yiyu Li, Youxin Jiang, Yuchen Gao, Yupeng Zhang, Zenan Zhou, and Zhiying Wu. 2023{\natexlab{a}}.
\newblock \href {https://doi.org/10.48550/ARXIV.2309.10305} {Baichuan 2: Open large-scale language models}.
\newblock \emph{CoRR}, abs/2309.10305.

\bibitem[{Yang et~al.(2023{\natexlab{b}})Yang, Li, Bing, and Lam}]{tft-once-opon-time}
Sen Yang, Xin Li, Lidong Bing, and Wai Lam. 2023{\natexlab{b}}.
\newblock \href {https://doi.org/10.18653/V1/2023.EMNLP-MAIN.728} {Once upon a time in graph: Relative-time pretraining for complex temporal reasoning}.
\newblock In \emph{Proceedings of the 2023 Conference on Empirical Methods in Natural Language Processing, {EMNLP} 2023, Singapore, December 6-10, 2023}, pages 11879--11895. Association for Computational Linguistics.

\bibitem[{Yu et~al.(2020)Yu, Jiang, Dong, and Feng}]{reclor}
Weihao Yu, Zihang Jiang, Yanfei Dong, and Jiashi Feng. 2020.
\newblock \href {https://openreview.net/forum?id=HJgJtT4tvB} {Reclor: {A} reading comprehension dataset requiring logical reasoning}.
\newblock In \emph{8th International Conference on Learning Representations, {ICLR} 2020, Addis Ababa, Ethiopia, April 26-30, 2020}. OpenReview.net.

\bibitem[{Zeng et~al.(2023)Zeng, Liu, Du, Wang, Lai, Ding, Yang, Xu, Zheng, Xia, Tam, Ma, Xue, Zhai, Chen, Liu, Zhang, Dong, and Tang}]{chatglm}
Aohan Zeng, Xiao Liu, Zhengxiao Du, Zihan Wang, Hanyu Lai, Ming Ding, Zhuoyi Yang, Yifan Xu, Wendi Zheng, Xiao Xia, Weng~Lam Tam, Zixuan Ma, Yufei Xue, Jidong Zhai, Wenguang Chen, Zhiyuan Liu, Peng Zhang, Yuxiao Dong, and Jie Tang. 2023.
\newblock \href {https://openreview.net/pdf?id=-Aw0rrrPUF} {{GLM-130B:} an open bilingual pre-trained model}.
\newblock In \emph{The Eleventh International Conference on Learning Representations, {ICLR} 2023, Kigali, Rwanda, May 1-5, 2023}. OpenReview.net.

\bibitem[{Zhang and Wan(2023)}]{situatedgen}
Yunxiang Zhang and Xiaojun Wan. 2023.
\newblock \href {http://papers.nips.cc/paper\_files/paper/2023/hash/d4f2bc9885ecbe30f65031819ef8699f-Abstract-Datasets\_and\_Benchmarks.html} {Situatedgen: Incorporating geographical and temporal contexts into generative commonsense reasoning}.
\newblock In \emph{Advances in Neural Information Processing Systems 36: Annual Conference on Neural Information Processing Systems 2023, NeurIPS 2023, New Orleans, LA, USA, December 10 - 16, 2023}.

\bibitem[{Zhang et~al.(2023)Zhang, Yao, Zhang, Tang, Ma, He, Wang, Gerstein, Wang, Liu, and Zhao}]{ignitingCot}
Zhuosheng Zhang, Yao Yao, Aston Zhang, Xiangru Tang, Xinbei Ma, Zhiwei He, Yiming Wang, Mark Gerstein, Rui Wang, Gongshen Liu, and Hai Zhao. 2023.
\newblock \href {https://doi.org/10.48550/ARXIV.2311.11797} {Igniting language intelligence: The hitchhiker's guide from chain-of-thought reasoning to language agents}.
\newblock \emph{CoRR}, abs/2311.11797.

\bibitem[{Zhao et~al.(2023)Zhao, Zhou, Li, Tang, Wang, Hou, Min, Zhang, Zhang, Dong, Du, Yang, Chen, Chen, Jiang, Ren, Li, Tang, Liu, Liu, Nie, and Wen}]{llmsurvey}
Wayne~Xin Zhao, Kun Zhou, Junyi Li, Tianyi Tang, Xiaolei Wang, Yupeng Hou, Yingqian Min, Beichen Zhang, Junjie Zhang, Zican Dong, Yifan Du, Chen Yang, Yushuo Chen, Zhipeng Chen, Jinhao Jiang, Ruiyang Ren, Yifan Li, Xinyu Tang, Zikang Liu, Peiyu Liu, Jian-Yun Nie, and Ji-Rong Wen. 2023.
\newblock \href {https://arxiv.org/abs/2303.18223} {A survey of large language models}.
\newblock \emph{Preprint}, arXiv:2303.18223.

\bibitem[{Zhou et~al.(2019)Zhou, Khashabi, Ning, and Roth}]{mctaco}
Ben Zhou, Daniel Khashabi, Qiang Ning, and Dan Roth. 2019.
\newblock \href {https://doi.org/10.18653/V1/D19-1332} {"going on a vacation" takes longer than "going for a walk": {A} study of temporal commonsense understanding}.
\newblock In \emph{Proceedings of the 2019 Conference on Empirical Methods in Natural Language Processing and the 9th International Joint Conference on Natural Language Processing, {EMNLP-IJCNLP} 2019, Hong Kong, China, November 3-7, 2019}, pages 3361--3367. Association for Computational Linguistics.

\bibitem[{Zhou et~al.(2021)Zhou, Richardson, Ning, Khot, Sabharwal, and Roth}]{tracie}
Ben Zhou, Kyle Richardson, Qiang Ning, Tushar Khot, Ashish Sabharwal, and Dan Roth. 2021.
\newblock \href {https://doi.org/10.18653/V1/2021.NAACL-MAIN.107} {Temporal reasoning on implicit events from distant supervision}.
\newblock In \emph{Proceedings of the 2021 Conference of the North American Chapter of the Association for Computational Linguistics: Human Language Technologies, {NAACL-HLT} 2021, Online, June 6-11, 2021}, pages 1361--1371. Association for Computational Linguistics.

\bibitem[{Zhu et~al.(2023)Zhu, Yang, Chen, Li, Lou, and Yang}]{emt-program}
Xinyu Zhu, Cheng Yang, Bei Chen, Siheng Li, Jian{-}Guang Lou, and Yujiu Yang. 2023.
\newblock \href {https://doi.org/10.18653/V1/2023.EMNLP-MAIN.787} {Question answering as programming for solving time-sensitive questions}.
\newblock In \emph{Proceedings of the 2023 Conference on Empirical Methods in Natural Language Processing, {EMNLP} 2023, Singapore, December 6-10, 2023}, pages 12775--12790. Association for Computational Linguistics.

\end{thebibliography}
\appendix
\section{\ours{} Details}
\label{appendix:a}
\begin{figure}[t]   
    \centering
    \includegraphics[clip, width=0.85\linewidth]{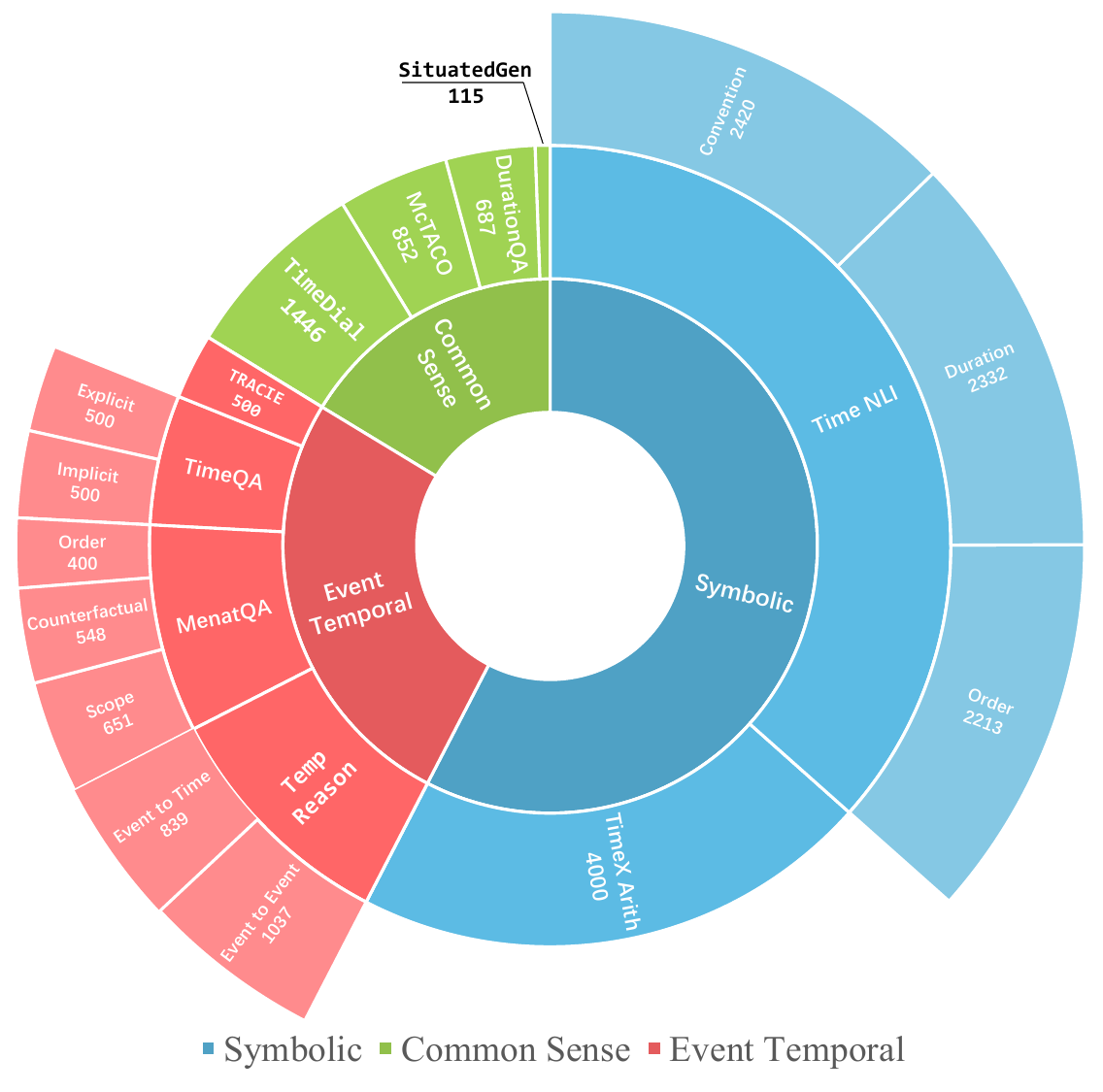}
    \caption{\label{fig:sunburst}
    The quantity and proportion of data for each task and its respective subtasks within \ours{}.
}
\end{figure}

\ours{} features 3 major categories, 10 tasks and 15 subtasks, each with distinct challenges, totaling 19,000 instances.
Detailed statistics are available in Figure~\ref{fig:sunburst} and Table~\ref{tab:dataset_challenges}.

\subsection{Benchmark Construction}
\label{appendix:a:benchmark_construction}
\paragraph{TimeX Arithmetic}\citep{tempreason}
TimeX Arithmetic data is derived from the \textit{l1: time-time} reasoning data in TempReason.
We retain 4,000 instances, where time expressions are calculated with a minimum unit of one day.

\paragraph{TimeX NLI}\citep{timexnli}
The original data of TimeXNLI is in NLI format, including three sub-tasks, 
\textit{Temp-Order}, \textit{Temp-Duration}, and \textit{Cross-Unit Duration},  including 6,140, 3,540, and 15,840 instances respectively.
We conduct a random sampling of 2,213, 2,332 and 2,429 entries, resulting in a combined total of 6,965 instances.

\paragraph{MCTACO}~\citep{mctaco}
The original MCTACO dataset consists of yes/no questions, containing 1,332 questions with 9,442 options.
To guarantee that the questions are presented in a 4-way multi-select style, we initially remove questions that have less than four options.
Subsequently, to ensure that each question has at least one correct option, we filter out questions where all options are labeled as "no".
For each remaining question, we randomly sample four options, striving to maintain a balance between correct and incorrect options.
In most cases, a question is accompanied by 2 correct and 2 incorrect options. 
A minority of questions have an option distribution of 1-3 or 3-1.
After the aforementioned filtering process, we obtain 852 pieces of data in a 4-way multi-select format.

\paragraph{DurationQA} ~\citep{durationqa}
The original DurationQA has the same format as MCTACO, which consists of 694 questions with 4,868 options.
Following the identical filtration procedure as MCTACO, we finally obtained a collection of 687 questions in a 4-way multi-select format.

\paragraph{TimeDial} ~\citep{timedial}
consists of 4-way multi-select instances in a two-person dialogue scenario.
We leave the original data unaltered and simply randomize the sequence of options, yielding 1,446 pieces of 4-way multi-select instances.

\paragraph{SituateGen}~\citep{situatedgen}
SituatedGen includes 1,220 test cases, which span across two distinct reasoning domains: \textit{time} and \textit{geography}.
We manually screen the original test data and retain those with clear time features for temporal reasoning evaluation, resulting in 115 instances.

\paragraph{TimeQA}~\citep{timeqa}
The original data of TimeQA includes two splits, \textit{Easy} and \textit{Hard}, with each question containing 20 Wikipedia paragraphs.
The excessively long context may exceed the model's maximum length limit and incur significant inference overhead. 
Therefore, we have reduced the context of the original data.
For the paragraphs in the original data, we refer to those containing the answer as relevant paragraphs, and the rest as irrelevant paragraphs.
For each question, we keep the first paragraph, all relevant paragraphs, and one random irrelevant paragraph as distractor.
This ensures that most questions have at least three paragraphs.
After that, we sample 500 pieces of data from those where the context length is less than 650 tokens.
For both \textit{Easy} and \textit{Hard} splits, we apply the aforementioned filtration, resulting in 500 questions each, totaling 1,000 instances.

\paragraph{TempReason}~\citep{tempreason}
TempReason dataset contains 5,397 entries for l2 (event-time reasoning) and 4,426 entries for l3 (event-event reasoning).
In the original dataset, each question corresponds with a text context and extracted facts.
Similar to TimeQA, we apply a filter based on context length. 
We preserve questions with a context length between 300 and 600 tokens, yielding 839 and 1,037 instances, respectively.
Notably, every remaining question is applicable to either context-based reasoning or fact-based reasoning.

\paragraph{MenatQA}~\citep{menatqa}
MenatQA consists of 999 data entries, formatted similarly to TimeQA, where each question is accompanied by several corresponding paragraphs.
Following the paper's proposed method, we modify the original data by incorporating the three time-sensitive factors: scope, order, and counterfactual.
Subsequently, for each factor, we randomly sample 400 instances, resulting in a total of 1,200 data points.

\paragraph{TRACIE}~\citep{tracie}
The original TRACIE dataset consists of yes/no type questions, containing 4,248 test instances.
We randomly sample 500 instances from the \textit{iid} split in the test set.

\subsection{Human Performance Evaluation}
\label{appendix:a:human_performance_evaluation}
Unless otherwise stated, the results of human evaluation are derived from original dataset papers. 
Please refer to the corresponding paper for human evaluation details.
TimeXNLI, Date Arith, and MCTACO are manually evaluated by three authors from the TimeBench team.
Within each subtask, we randomly sample 50 instances, and the average of the performances by three human evaluators is considered the final human performance.

\subsection{Task Formats}
\label{appendix:a:task_formats}
\ours{} is a multispectral benchmark, which features four different task formats.

\paragraph{Multi-Select Questions}
Previous work utilizes the Multiple Choice (MC) form, which requires models to select the only correct answer from the options. However, this task form has shortcuts and may not truly reflect the model's abilities. 
To address this, we employ the Multi-Select (M-S) task form, where the model needs to select all possible correct answers from the options provided.
In our task, each question presents four options, with at most two of them being correct.

\paragraph{Natural Language Inference} is the task of determining the logical relationship between two pieces of text. 
Specifically, given a premise and a hypothesis, the model needs to determine whether the hypothesis can be inferred from the premise and output entailment, contradiction, or neutral.
Our tasks focus on the entailment in temporal domains.

\paragraph{Free-form Reading Comprehension} requires models to answer questions based on the provided context, and the ground truth answer is free-form without pre-defined format restrictions.

\paragraph{Constrained Text Generation}
refers to the task of generating text under certain constraints.
The task is keyword-constrained text generation, where the model takes keywords as input and outputs sentences that include those keywords.

\subsection{Evaluation Metrics}
\label{appendix:a:evaluation_metrics}
Accuracy is used for NLI and date arithmetic tasks.
M-S tasks are evaluated using option-level EM and F1.
FRC tasks (excluding date arithmetic) are assessed with token-level EM and F1.
For CTG task, we take the average of multiple generation metrics, which are outlined as follows.

\paragraph{Metrics for SituatedGen}
Following SituatedGen~\citep{situatedgen}, we use BLEU-4~\citep{bleu}, METEOR~\citep{meteor}, ROUGE-L~\citep{rouge}, CIDEr~\citep{cider}, and MATCH~\citep{situatedgen} scores to metric the results of CTG.\footnote{We utilize \href{https://github.com/salaniz/pycocoevalcap
}{\textsf{pycocoevalcap}} package to calucate BLEU-4, METEOR, ROUGE-L, CIDEr.}

The overall score is calculated as the sum of the above scores. We set the weight of CIDEr to $1/10$ for balancing when summation.
\begin{multline*}
    S = \textrm{BLEU-4} + \textrm{METEOR} + \textrm{ROUGE-L} \\
    + \textrm{CIDER} / 10 + \textrm{MATCH}
\end{multline*}

As the overall score $S$ does not represent a percentile, we proceed to normalize the models' scores to align with humans' relative performance levels.

\section{Supplemental Materials}

\subsection{Models}
\label{appendix:models}

\paragraph{ChatGPT-3.5/GPT-4}\citep{gpt3.5,gpt4}
ChatGPT is a chat model aligned through SFT and RLHF based on GPT-3~\citep{gpt3}. 
GPT-4 is an upgraded version of ChatGPT with enhanced reasoning capabilities, making it the most powerful LLM.
Unless otherwise stated, ChatGPT refers to \textit{gpt-3.5-turbo-0613} and GPT-4 refers to \textit{gpt-4-0613}.

\paragraph{Llama2/Vicuna-1.5}\citep{llama2,vicuna:chiang}
LLaMA2 is an open foundation model trained on 2T tokens with efficient grouped-query attention~\cite{gqa}.
LLaMA2-chat is the official aligned model with SFT and RLHF, and 
Vicuna-1.5 is aligned with SFT only by the community\footnote{\url{https://lmsys.org/}}.

\paragraph{Baichuan2}\citep{baichuan2} is an open foundation model pre-trained on 2.6T tokens, which is competitive with LLaMA2.
Baichuan2-chat is the official aligned model with SFT and RLHF.

\paragraph{Mistral}\citep{mistral} is a 7B open foundation model incorporating efficient grouped-query attention~\citep{gqa} and sliding windows attention~\citep{swa}.
It achieves the strongest performance among models of its size, even surpassing LLaMA2-13B.
Mistral-instruct is the officially aligned model with SFT only.

\paragraph{ChatGLM3}\citep{chatglm} is an open-source bilingual LLM for Chinese and English, exhibiting competitive performance under 10B.

\paragraph{FLAN-T5}\citep{flant5} is an open-source instruction model built on top of T5~\citep{t5} through instruction fine-tuning.

\subsection{Full Results}
\label{appendix:full_results}
The overall score is derived from the average of all corresponding metrics.
For brevity, we omit some F1 scores in tables in the main text.
Please refer to Table~\ref{tab:full_results} for the full experimental results.
The full results of SituatedGen can be found in Table~\ref{tab:full_sitgen}.

\subsection{Prompts}
\label{appendix:prompts}
The prompt formats are showcased in Figure~\ref{fig:prompt-instruction}.
The demonstrations can be found from Figure~\ref{fig:prompt-demos-timexnli} to~\ref{fig:prompt-demos-tempreason-ee}.

\begin{figure*}[t]
    \centering
    \includegraphics[clip, width=\linewidth]{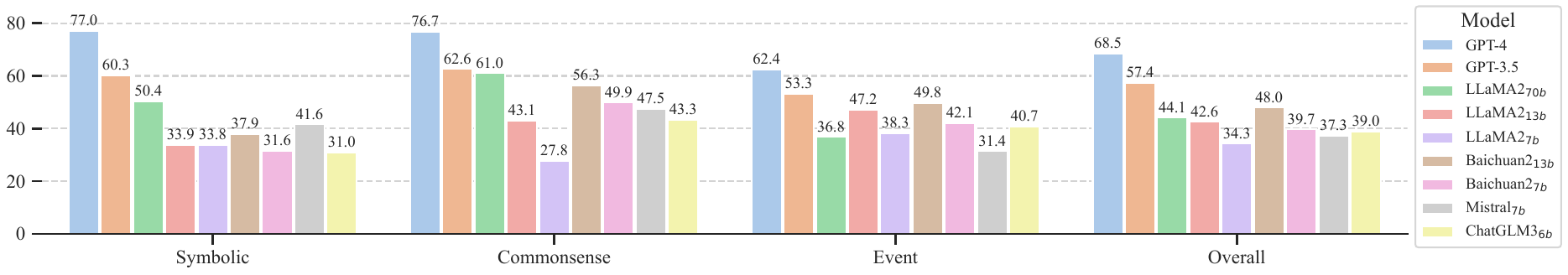}
    \includegraphics[clip, width=\linewidth]{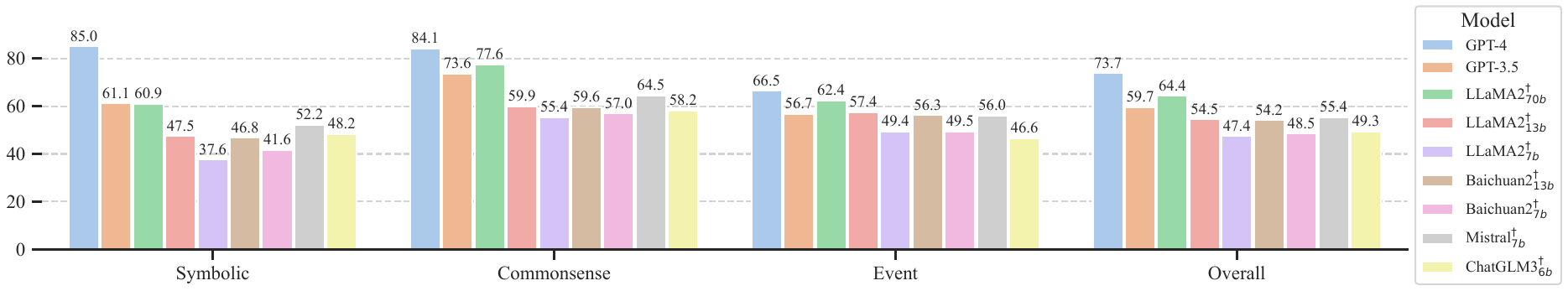}
    \caption{\label{fig:overall}
    Performance comparison between state-of-the-art LLMs. \textbf{Up}: GPT-4/3.5 and alignment models under zero-shot setting. \textbf{Down}: GPT-4/3.5 and base models under few-shot setting.  
}
\end{figure*}

\begin{table*}[t]
\centering
\setlength\tabcolsep{2pt}
\begin{tabular}{l|ccc}
\toprule
\textbf{Dataset} & \textbf{Format} & \textbf{\#} & \textbf{Challenges} \\
\midrule
\multicolumn{4}{l}{\textbf{Symbolic}} \\
\midrule
TimeX Arith & FRC & 4,000 & TimeX Arithmetic \\
TimeX NLI & NLI & 6,965 & TimeX Causality \\
\rowcolor{Gray}
~~~- \textit{Order} &- & 2,213 & order \\ 
\rowcolor{Gray}
~~~- \textit{Duration} &- & 2,332 & duration \\
\rowcolor{Gray}
~~~- \textit{Conversion} &- &  2,420 & duration + time unit conversion \\ 
\midrule
\multicolumn{4}{l}{\textbf{Commonsense}} \\
\midrule
MCTACO & M-S & 852 & Temporal Commonsense \\
TimeDial & M-S & 1,446 & Temporal Commonsense \\
DurationQA & M-S & 687 & Event Duration \\
SituatedGen & CTG & 115 & Temporal Commonsense \\
\midrule
\multicolumn{4}{l}{\textbf{Event}} \\
\midrule
TimeQA & FRC & 1,000 & Context-based Reasoning \\
\rowcolor{Gray}
~~~- \textit{Explicit} &- & 500 & explicit, event-time reasoning \\
\rowcolor{Gray}
~~~- \textit{Implicit} &- & 500 & implicit, event-time reasoning \\
MenatQA & FRC & 1,599 & Implicit, Context-based Reasoning \\
\rowcolor{Gray}
~~~- \textit{Order} &- & 400 & event-time reasoning \\
\rowcolor{Gray}
~~~- \textit{Scope} &- & 400 & event-time reasoning \\
\rowcolor{Gray}
~~~- \textit{Counterfactual} &- & 400 & event-time reasoning \\
TempReason & FRC & 1,876 & Implicit, Fact-based Reasoning \\
\rowcolor{Gray}
~~~- \textit{l2 (e2t)} &- & 839 & event-time reasoning \\
\rowcolor{Gray}
~~~- \textit{l3 (e2e)} &- & 1,037 & event-event reasoning \\
TRACIE & NLI & 500 & Implicit, Implied Event-Event Reasoning \\
\midrule
\textit{In total} &  & 19,000 & \\
\bottomrule
\end{tabular}
\caption{The statistics, task formats and challenges in \ours{}.}
\label{tab:dataset_challenges}
\end{table*}

\begin{table*}[t]
\centering
\begin{tabular}{l|ccccc|cc}
\toprule
Method & BLEU-4 & METEOR & ROUGE-L & CIDEr & MATCH & Overall & Norm \\
\midrule
Human & 39.9 & 40.4 & 56.3 & 397 & 98.1 & 274.4 & 100.0 \\
\midrule
GPT-4 & 8.23 & 31.27 & 28.84 & 38.45 & \textbf{90.41} & 162.59 & 59.25 \\
\addMethod{FS} & \textbf{28.64} & \textbf{38.99} & \textbf{55.69} & \textbf{298.64} & \textbf{90.11} & \textbf{243.29} & \textbf{88.66} \\
\midrule
GPT-3.5 & 13.38 & 30.12 & 35.91 & 125.41 & \textbf{78.76} & 170.70 & 62.21 \\
\addMethod{FS} & \textbf{27.24} & \textbf{33.77} & \textbf{51.18} & \textbf{282.75} & 76.54 & \textbf{217.01} & \textbf{79.08} \\
\midrule
LLaMA2$_{70b}$ & 5.15 & 13.62 & 15.83 & 22.07 & 31.79 & 68.60 & 25.00 \\
\addMethod{FS} & 19.10 & 29.09 & 41.74 & 171.36 & 65.29 & 172.35 & 62.81 \\
\midrule
LLaMA2$_{13b}$ & 4.66 & 21.43 & 20.80 & 17.72 & 61.62 & 110.28 & 40.19 \\
\addMethod{FS} & 15.15 & 27.49 & 37.55 & 138.13 & 64.94 & 158.93 & 57.92 \\
\midrule
LLaMA2$_{7b}$ & 2.77 & 13.46 & 14.69 & 14.34 & 34.83 & 67.18 & 24.48 \\
\addMethod{FS} & 6.90 & 15.82 & 21.77 & 52.99 & 33.81 & 83.60 & 30.47 \\
\midrule
Baichuan2$_{13b}$ & 8.33 & 25.86 & 30.07 & 82.63 & 70.63 & 143.15 & 52.17 \\
\addMethod{FS} & 15.79 & 30.23 & 40.96 & 169.14 & 71.01 & 174.91 & 63.74 \\
\midrule
Baichuan2$_{7b}$ & 5.17 & 21.99 & 23.73 & 44.80 & 59.85 & 115.22 & 41.99 \\
\addMethod{FS} & 15.06 & 23.45 & 32.29 & 137.94 & 52.04 & 136.64 & 49.79 \\
\midrule
Vicuna1.5$_{13b}$ & 7.73 & 26.35 & 29.15 & 69.16 & 71.91 & 142.06 & 51.77 \\
\addMethod{FS} & 6.85 & 18.66 & 25.99 & 92.96 & 46.19 & 106.99 & 38.99 \\
\midrule
Vicuna1.5$_{7b}$ & 6.29 & 24.34 & 26.91 & 46.90 & 68.84 & 131.07 & 47.77 \\
\addMethod{FS} & 20.71 & 30.19 & 45.20 & 203.20 & 67.58 & 184.00 & 67.05 \\
\midrule
FLAN-T5 & 16.20 & 24.43 & 29.38 & 95.17 & 56.38 & 135.91 & 49.53 \\
\addMethod{FS} & 12.88 & 30.38 & 36.27 & 92.20 & 76.44 & 165.19 & 60.20 \\
\midrule
Mistral$_{7b}$ & 5.82 & 22.89 & 24.19 & 44.03 & 63.74 & 121.03 & 44.11 \\
\addMethod{FS} & 18.96 & 29.02 & 43.15 & 185.61 & 63.24 & 172.93 & 63.02 \\
\midrule
ChatGLM3$_{6b}$ & 6.56 & 21.11 & 21.96 & 41.48 & 53.02 & 106.80 & 38.92 \\
\addMethod{FS} & 10.53 & 24.17 & 33.44 & 124.50 & 56.94 & 137.53 & 50.12 \\
\midrule
\midrule
\rowcolor{Gray}
LLaMA2$^\dag_{70b}$ & \textbf{22.34} & \textbf{33.03} & \textbf{50.93} & \textbf{243.31} & 74.96 & \textbf{205.59} & \textbf{74.92} \\ 
\rowcolor{Gray}
LLaMA2$^\dag_{13b}$ & 17.54 & 29.44 & 45.21 & 200.14 & 65.64 & 177.84 & 64.81 \\
\rowcolor{Gray}
LLaMA2$^\dag_{7b}$ & 17.49 & 28.33 & 45.24 & 202.08 & 59.98 & 171.25 & 62.41 \\
\rowcolor{Gray}
Baichuan2$^\dag_{13b}$ & 17.86 & 29.75 & 44.28 & 198.83 & 66.35 & 178.12 & 64.91 \\
\rowcolor{Gray}
Baichuan2$^\dag_{7b}$ & 15.30 & 27.54 & 41.80 & 171.59 & 62.40 & 164.20 & 59.84 \\
\rowcolor{Gray}
Mistral$^\dag_{7b}$ & 14.54 & 27.39 & 41.72 & 168.89 & 59.42 & 159.96 & 58.30 \\
\rowcolor{Gray}
ChatGLM3$^\dag_{6b}$ & 17.11 & 29.35 & 40.74 & 156.49 & 66.18 & 169.02 & 61.60 \\
\bottomrule
\end{tabular}
\caption{Full results of SituatedGen. Aligned models are under zero-shot setting by default. The top-3 results are \textbf{bold}. Methods with \dag ~are base models without alignment, under few-shot setting. We consider human performance as 100 points and normalize models' results accordingly.}
\label{tab:full_sitgen}
\end{table*}
\clearpage
\onecolumn
\begin{landscape}
\centering
\scriptsize
\setlength\tabcolsep{2pt}
\begin{longtable}[c]{l|cccc|ccccccc|ccccccccccccccc|cccc}
\toprule

\multirow{3.5}{*}{\textbf{Method}} & \multicolumn{4}{c|}{\textbf{Symbolic}} & \multicolumn{7}{c|}{\textbf{Commonsense}} & \multicolumn{15}{c|}{\textbf{Event}} & \multicolumn{4}{c}{\textbf{Overall}} \\
\cmidrule(lr){2-5}\cmidrule(lr){6-12}\cmidrule(lr){13-27}\cmidrule(lr){28-31}

 & \multicolumn{3}{c}{TimeXNLI} & Date Arith & \multicolumn{2}{c}{DurationQA} & \multicolumn{2}{c}{McTACO} & \multicolumn{2}{c}{TimeDial} & SitGen & \multicolumn{4}{c}{TimeQA} & \multicolumn{6}{c}{MenatQA} & \multicolumn{4}{c}{TempReason} & TRACIE & \multirow{2}{*}{Sym.} & \multirow{2}{*}{Comm.} & \multirow{2}{*}{Event} & \multirow{2}{*}{Avg.} \\
 & \textit{s1} & \textit{s2} & \textit{s3} & \textit{Acc} & \textit{EM} & \textit{F1} & \textit{EM} & \textit{F1} & \textit{EM} & \textit{F1} & \textit{Norm} & \textit{E-EM} & \textit{E-F1} & \textit{H-EM} & \textit{H-F1} & \textit{S-EM} & \textit{S-F1} & \textit{O-EM} & \textit{O-F1} & \textit{C-EM} & \textit{C-F1} & \textit{L2-EM} & \textit{L2-F1} & \textit{L3-EM} & \textit{L3-F1} & \textit{Acc} & & & & \\
\endhead

\midrule
Human & 98.0 & 96.0 & 92.0 & 100.0 & 64.0 & 80.8 & 75.8 & 87.1 & 97.8 & 97.8 & 100.0 & 89.0 & 93.3 & 87.0 & 91.1 & 82.0 & 85.6 & 84.0 & 87.3 & 76.0 & 79.9 & 96.0 & 97.1 & 94.0 & 95.3 & 82.5 & 96.5 & 91.4 & 89.0 & 91.5 \\
\midrule

GPT-4 & 78.6 & 76.0 & 50.7 & \textbf{98.0} & 35.0 & 59.2 & 61.2 & 80.0 & 72.0 & 91.1 & 59.3 & 48.9 & 60.6 & 40.4 & 46.5 & 44.4 & 57.0 & 49.0 & 57.0 & 22.0 & 23.1 & 91.0 & 95.3 & 94.0 & 95.0 & 64.8 & 75.8 & 72.4 & 62.4 & 68.3 \\
\addMethod{CoT} & 80.0 & 76.0 & 60.0 & 92.0 & 35.0 & 58.1 & 67.0 & 82.6 & 65.0 & 89.3 & - & 50.0 & 61.3 & 33.0 & 41.2 & 43.4 & 54.6 & \textbf{53.0} & \textbf{59.6} & 20.0 & 22.6 & \textbf{93.0} & \textbf{97.0} & 93.0 & 94.5 & 58.0 & 77.0 & 76.7 & 61.1 & 68.5 \\
\addMethod{FS} & 85.3 & 73.3 & 53.3 & 100.0 & \textbf{51.0} & \textbf{64.8} & \textbf{77.6} & \textbf{88.3} & \textbf{85.0} & \textbf{94.6} & \textbf{88.6} & \textbf{59.2} & \textbf{73.7} & 40.0 & 51.0 & \textbf{59.6} & \textbf{72.4} & 48.0 & 54.8 & \textbf{25.3} & \textbf{28.7} & 86.0 & 92.4 & \textbf{94.8} & \textbf{95.9} & 62.8 & 78.0 & \textbf{84.1} & \textbf{66.5} & \textbf{73.7} \\
\addMethod{FS CoT} & \textbf{92.0} & \textbf{84.0} & \textbf{64.0} & 100.0 & 42.0 & 55.1 & 68.0 & 72.3 & 79.0 & 93.4 & - & 48.0 & 66.9 & \textbf{44.4} & \textbf{52.8} & 48.5 & 65.3 & 44.0 & 52.6 & 22.0 & 25.9 & 91.0 & 96.9 & 93.0 & 94.6 & \textbf{66.4} & \textbf{85.0} & 73.6 & 65.2 & 72.1 \\
\midrule

GPT-3.5 & 45.4 & 67.6 & 31.2 & \textbf{97.0} & 19.2 & 50.5 & 34.1 & 68.6 & 39.2 & 69.1 & 62.3 & \textbf{60.5} & \textbf{70.8} & 29.5 & 35.4 & 36.5 & 40.9 & 37.5 & 43.9 & 21.0 & 22.9 & 73.6 & 81.2 & 61.8 & 73.8 & \textbf{57.4} & 60.3 & 62.6 & 53.3 & 57.4 \\
\addMethod{CoT} & 33.6 & 64.8 & 33.6 & 71.0 & 12.4 & 23.2 & 28.1 & 45.1 & 34.6 & 67.0 & - & 52.5 & 64.4 & 29.0 & 35.1 & 35.8 & 39.7 & 38.5 & 42.9 & 24.0 & 26.3 & 32.0 & 57.6 & 54.2 & 68.1 & 52.0 & 50.8 & 45.1 & 48.3 & 48.3 \\
\addMethod{FS} & 52.0 & 68.4 & 31.6 & 63.6 & \textbf{42.8} & \textbf{67.7} & \textbf{43.5} & \textbf{71.2} & 47.8 & \textbf{76.4} & \textbf{79.1} & 53.8 & 66.1 & \textbf{37.9} & \textbf{48.4} & 37.8 & \textbf{43.2} & \textbf{43.5} & \textbf{51.6} & 16.0 & 17.9 & 77.7 & 84.7 & \textbf{70.0} & \textbf{78.0} & 55.0 & 53.9 & \textbf{73.6} & 55.6 & \textbf{59.7} \\
\addMethod{FS CoT} & \textbf{51.6} & \textbf{71.8} & \textbf{36.6} & 84.4 & 20.8 & 41.2 & 21.4 & 38.1 & \textbf{48.3} & 71.1 & - & 56.5 & 68.0 & 37.5 & 47.0 & \textbf{38.1} & 42.5 & 37.5 & 41.7 & \textbf{33.0} & \textbf{37.8} & \textbf{86.2} & \textbf{89.9} & 68.0 & 76.6 & 50.2 & \textbf{61.1} & 50.1 & \textbf{56.7} & 56.6 \\
\midrule

LLaMA2$_{70b}$ & 44.0 & 47.0 & 32.0 & \textbf{78.5} & \textbf{12.7} & 59.2 & \textbf{23.0} & \textbf{68.9} & \textbf{10.0} & 57.0 & 25.0 & 28.0 & 40.8 & 31.0 & 40.6 & 8.0 & 18.9 & 11.0 & 16.6 & 9.0 & 12.0 & 50.0 & 63.5 & 39.0 & 54.5 & 48.0 & 50.4 & 52.5 & 36.8 & 44.1 \\
\addMethod{CoT} & 30.0 & \textbf{66.0} & 28.0 & 53.5 & 8.0 & 57.3 & 21.0 & 67.1 & 9.0 & \textbf{58.6} &  & 17.0 & 31.4 & 13.0 & 19.5 & 5.0 & 12.2 & 8.0 & 12.7 & \textbf{18.0} & 20.8 & 12.0 & 37.5 & 20.0 & 40.5 & 51.0 & 44.4 & 61.0 & 28.2 & 39.1 \\
\addMethod{FS} & 49.0 & 42.0 & 38.0 & 62.0 & 1.3 & \textbf{61.2} & 13.0 & 66.5 & 6.0 & 56.6 & \textbf{62.8} & \textbf{41.0} & \textbf{51.1} & 16.0 & 20.0 & 8.0 & 16.4 & 17.0 & 19.9 & \textbf{18.0} & 18.7 & 34.0 & 52.2 & 31.0 & 41.1 & 51.0 & 47.8 & \textbf{61.8} & 33.8 & 44.3 \\
\addMethod{FS CoT} & 54.0 & 63.0 & \textbf{40.0} & 69.5 & 8.0 & 55.2 & 21.5 & 62.1 & 6.0 & 56.4 &  & 36.6 & 50.9 & \textbf{34.0} & \textbf{42.4} & \textbf{28.0} & \textbf{38.6} & \textbf{19.0} & \textbf{29.3} & \textbf{18.0} & \textbf{21.9} & \textbf{77.0} & \textbf{83.1} & \textbf{65.0} & \textbf{74.7} & \textbf{57.0} & \textbf{56.6} & 57.9 & \textbf{49.7} & \textbf{53.2} \\
\midrule

LLaMA2$_{13b}$ & 30.0 & 49.0 & 34.0 & \textbf{22.5} & 4.0 & 38.5 & 8.5 & 40.6 & 10.0 & 35.4 & 57.9 & \textbf{46.0} & \textbf{61.9} & 21.0 & 30.5 & 28.0 & 46.1 & \textbf{23.0} & 36.1 & 18.0 & 26.9 & 43.0 & 53.1 & 55.0 & \textbf{69.4} & 49.0 & 33.9 & 43.1 & 46.6 & 42.6 \\
\addMethod{CoT} & 36.0 & 50.0 & 38.0 & 6.0 & 7.3 & 39.2 & \textbf{14.0} & 51.7 & 10.0 & 36.9 & - & 45.0 & 58.7 & \textbf{30.0} & \textbf{38.9} & 20.0 & 40.9 & 18.0 & 32.5 & 21.0 & \textbf{33.6} & 43.0 & 58.0 & \textbf{56.0} & 68.4 & 47.0 & 32.5 & 42.6 & \textbf{47.3} & 42.4 \\
\addMethod{FS} & \textbf{43.0} & \textbf{57.0} & \textbf{60.0} & 20.5 & 9.0 & 46.8 & 8.0 & \textbf{66.6} & \textbf{15.0} & \textbf{62.3} & \textbf{40.2} & 24.0 & 34.2 & 17.0 & 18.4 & 11.0 & 25.9 & 5.0 & 14.6 & \textbf{22.0} & 33.3 & 54.0 & 68.1 & 50.0 & 64.8 & 47.0 & \textbf{45.1} & \textbf{54.0} & 38.3 & 43.9 \\
\addMethod{FS CoT} & 37.0 & 55.0 & 50.0 & 33.0 & \textbf{12.0} & \textbf{49.5} & 11.0 & 45.6 & 8.0 & 44.5 & - & 35.0 & 46.0 & 21.0 & 25.4 & \textbf{34.0} & \textbf{46.7} & 23.0 & \textbf{36.5} & 7.0 & 16.5 & \textbf{72.0} & \textbf{80.8} & 54.0 & 66.2 & \textbf{50.0} & 43.8 & 46.5 & 46.0 & \textbf{45.5} \\
\midrule

LLaMA2$_{7b}$ & 39.0 & 53.0 & 30.0 & \textbf{13.0} & 2.7 & 39.3 & 4.0 & 41.0 & 1.0 & 6.3 & 24.5 & \textbf{37.0} & 49.0 & 14.0 & 29.0 & 7.0 & 26.8 & 8.0 & 21.1 & \textbf{9.0} & 16.0 & \textbf{48.0} & \textbf{63.9} & \textbf{32.0} & 47.9 & 49.0 & 33.8 & 27.8 & 37.8 & 34.3 \\
\addMethod{CoT} & \textbf{44.0} & 50.0 & 33.0 & 5.0 & 2.7 & 35.0 & 4.5 & 40.0 & 1.0 & 1.7 & - & 27.0 & 49.9 & 17.0 & 31.6 & \textbf{11.0} & \textbf{31.4} & \textbf{10.0} & \textbf{24.5} & 7.0 & 17.8 & 44.0 & 56.9 & \textbf{32.0} & \textbf{48.1} & 46.0 & 33.0 & 25.6 & \textbf{38.3} & 34.3 \\
\addMethod{FS} & \textbf{44.0} & \textbf{60.0} & 34.0 & 11.0 & 4.0 & \textbf{62.8} & 8.0 & 64.7 & 8.0 & 40.0 & \textbf{30.5} & 36.0 & 50.8 & 20.0 & 29.4 & 5.0 & 22.3 & 6.0 & 18.0 & 6.0 & \textbf{17.9} & 12.0 & 36.3 & 23.0 & 44.3 & \textbf{53.0} & \textbf{37.3} & 49.5 & 34.0 & \textbf{38.7} \\
\addMethod{FS CoT} & 38.0 & 51.0 & \textbf{36.0} & 14.5 & \textbf{11.0} & 42.8 & \textbf{25.0} & \textbf{65.6} & \textbf{13.0} & \textbf{53.4} & \textbf{-} & 36.0 & \textbf{53.5} & \textbf{21.0} & \textbf{34.1} & 1.0 & 13.6 & 3.0 & 11.2 & 5.0 & 14.0 & 22.0 & 46.7 & 21.0 & 42.3 & 51.0 & 34.9 & \textbf{53.9} & 33.3 & 37.8 \\
\midrule

Baichuan2$_{13b}$ & 41.0 & \textbf{61.0} & 37.0 & \textbf{12.5} & 4.0 & 52.0 & 18.5 & 63.4 & 15.0 & 57.7 & 52.2 & 45.0 & 55.4 & 29.0 & 34.6 & 31.0 & 48.8 & \textbf{34.0} & \textbf{44.3} & 30.0 & 39.5 & 40.0 & 57.4 & 45.0 & 61.4 & 49.0 & 37.9 & 56.3 & 48.8 & 48.0 \\
\addMethod{CoT} & 40.0 & 57.0 & 31.0 & 10.0 & 3.3 & 44.6 & 20.0 & 61.9 & 13.0 & 58.1 & - & 36.0 & 41.5 & \textbf{36.0} & 40.9 & \textbf{39.0} & \textbf{52.0} & 27.0 & 38.5 & 29.0 & \textbf{43.2} & 46.0 & 62.8 & 46.0 & \textbf{64.3} & \textbf{55.0} & 34.5 & 54.9 & 49.8 & 46.7 \\
\addMethod{FS} & 43.0 & 59.0 & 40.0 & 42.5 & \textbf{24.7} & \textbf{62.1} & \textbf{27.5} & \textbf{70.2} & \textbf{18.0} & \textbf{58.9} & \textbf{63.7} & \textbf{47.0} & \textbf{60.7} & 35.0 & \textbf{45.7} & 37.0 & 51.9 & 31.0 & 41.5 & 19.0 & 31.8 & \textbf{73.0} & \textbf{81.1} & \textbf{48.0} & 59.4 & 48.0 & 46.1 & \textbf{63.7} & \textbf{52.5} & \textbf{53.7} \\
\addMethod{FS CoT} & \textbf{45.0} & 54.0 & \textbf{48.0} & 47.0 & 10.7 & 44.4 & 27.0 & 68.8 & 15.0 & 55.0 & - & 43.0 & 57.8 & 27.0 & 36.7 & 38.0 & 49.8 & 34.0 & 40.7 & \textbf{33.0} & 43.0 & 72.8 & 80.4 & 43.0 & 60.2 & 44.0 & \textbf{48.5} & 56.1 & 51.6 & 51.7 \\
\midrule

Baichuan2$_{7b}$ & 35.0 & \textbf{50.0} & 37.0 & \textbf{4.5} & 4.0 & 47.9 & 10.5 & 55.3 & 15.0 & 54.3 & 42.0 & 26.0 & 41.5 & 20.0 & 34.7 & 20.0 & 35.2 & 19.0 & 31.2 & 6.0 & 20.4 & 22.0 & 43.4 & 29.0 & 47.7 & \textbf{55.0} & 31.6 & \textbf{49.9} & 38.6 & 39.7 \\
\addMethod{CoT} & 38.0 & 43.0 & 32.0 & 1.0 & 5.3 & 37.9 & 13.0 & 58.0 & 15.0 & 44.2 & - & 41.0 & 53.5 & 28.0 & 38.8 & 29.0 & 39.9 & 23.0 & 33.2 & 18.0 & 29.3 & 21.0 & 41.2 & 29.0 & 47.2 & 54.0 & 28.5 & 46.7 & 42.1 & 39.4 \\
\addMethod{FS} & 40.0 & \textbf{50.0} & 36.0 & 20.0 & \textbf{28.7} & \textbf{59.4} & \textbf{26.5} & \textbf{66.9} & \textbf{17.0} & \textbf{53.0} & \textbf{49.8} & \textbf{45.0} & \textbf{60.7} & \textbf{30.0} & \textbf{42.1} & 27.0 & 37.8 & 23.0 & 35.7 & 10.0 & 20.4 & 40.0 & 57.4 & 37.0 & 53.0 & 51.0 & 36.5 & 57.3 & 44.8 & 45.8 \\
\addMethod{FS CoT} & \textbf{41.0} & \textbf{50.0} & 36.0 & 23.5 & 13.0 & 45.7 & 17.5 & 58.1 & 7.0 & 39.2 & - & 36.0 & 51.2 & 29.0 & \textbf{43.0} & \textbf{42.0} & \textbf{52.5} & \textbf{25.0} & \textbf{39.3} & \textbf{20.0} & \textbf{31.0} & \textbf{57.0} & \textbf{70.1} & \textbf{39.0} & \textbf{60.2} & 49.0 & \textbf{37.6} & 47.7 & \textbf{49.5} & \textbf{46.0} \\
\midrule

Vicuna1.5$_{13b}$ & 35.0 & 50.0 & 36.0 & \textbf{15.0} & 8.0 & 39.2 & 21.5 & 59.1 & 7.0 & 34.2 & \textbf{51.8} & 43.0 & 60.4 & 29.0 & 37.0 & 38.0 & 46.8 & 22.0 & 37.4 & 17.0 & 23.2 & 14.0 & 42.1 & 13.0 & 43.6 & 46.0 & 34.0 & \textbf{46.1} & 42.1 & 41.1 \\
\addMethod{CoT} & 42.0 & 51.0 & 37.0 & 3.0 & 1.3 & 29.8 & 11.5 & 50.0 & 7.0 & 33.7 & - & 44.0 & 56.9 & 31.0 & 36.4 & 16.0 & 38.2 & 25.0 & 37.7 & 13.0 & 20.4 & 31.0 & 49.0 & 29.0 & 49.1 & 51.0 & 33.3 & 37.8 & 42.3 & 39.0 \\
\addMethod{FS} & \textbf{48.0} & 57.0 & 38.0 & 30.5 & 7.3 & 33.6 & \textbf{27.5} & \textbf{57.0} & \textbf{13.0} & 40.3 & 39.0 & 45.0 & 58.3 & 23.0 & 25.9 & 38.0 & 42.6 & 26.0 & \textbf{41.4} & 18.0 & 20.1 & 51.0 & 61.8 & 28.0 & 42.6 & \textbf{56.0} & 43.4 & 42.5 & 43.6 & 43.3 \\
\addMethod{FS CoT} & 38.0 & \textbf{59.0} & \textbf{39.0} & 39.5 & \textbf{10.7} & \textbf{37.4} & 14.0 & 45.8 & 12.0 & \textbf{41.6} & \textbf{-} & \textbf{47.0} & \textbf{59.5} & \textbf{27.0} & \textbf{30.7} & \textbf{39.0} & \textbf{48.1} & \textbf{31.0} & 35.9 & \textbf{26.0} & \textbf{31.2} & \textbf{71.0} & \textbf{77.5} & \textbf{53.0} & \textbf{65.5} & 52.0 & \textbf{43.9} & 41.6 & \textbf{50.1} & \textbf{46.7} \\
\midrule

Vicuna1.5$_{7b}$ & 37.0 & \textbf{58.0} & \textbf{43.0} & \textbf{5.0} & 1.3 & 40.4 & \textbf{9.5} & \textbf{52.5} & 6.0 & 32.0 & 47.8 & \textbf{35.0} & \textbf{47.1} & 11.0 & 18.5 & \textbf{20.0} & \textbf{35.7} & 15.0 & 25.7 & 12.0 & 17.3 & 14.0 & 33.0 & 14.0 & 46.8 & \textbf{54.0} & 35.8 & 43.2 & 34.8 & \textbf{37.1} \\
\addMethod{CoT} & 36.0 & 50.0 & 36.0 & 1.5 & 1.3 & 39.4 & 8.5 & 49.2 & 9.0 & 36.2 & - & 30.0 & 40.9 & \textbf{14.0} & \textbf{24.6} & 16.0 & 26.2 & 14.0 & \textbf{28.5} & 12.0 & \textbf{25.0} & 9.0 & 27.7 & 7.0 & 40.3 & \textbf{54.0} & 30.9 & 41.6 & 33.4 & 34.4 \\
\addMethod{FS} & \textbf{43.0} & 57.0 & 37.0 & 8.5 & \textbf{3.3} & \textbf{44.6} & 5.5 & 42.1 & 7.0 & 36.8 & \textbf{67.1} & 24.0 & 31.9 & 12.0 & 14.9 & 16.0 & 21.8 & \textbf{20.0} & 27.5 & \textbf{17.0} & 22.2 & 13.0 & 34.3 & 6.0 & 32.2 & \textbf{54.0} & \textbf{36.4} & \textbf{47.7} & 29.9 & 35.9 \\
\addMethod{FS CoT} & 35.0 & 54.0 & 35.0 & 8.0 & 2.7 & 37.2 & 10.0 & 47.5 & \textbf{10.0} & \textbf{41.3} & \textbf{-} & 31.0 & 39.9 & 13.0 & 16.6 & 15.0 & 26.7 & 15.0 & 23.8 & 16.0 & 23.1 & \textbf{55.0} & \textbf{66.3} & \textbf{32.0} & \textbf{48.1} & 43.0 & 33.0 & 42.0 & \textbf{35.9} & 36.4 \\
\midrule

FLANT5$_{11b}$ & 53.0 & 63.0 & 43.0 & 0.0 & 4.0 & \textbf{52.0} & 14.0 & 65.0 & 13.0 & 47.7 & 49.5 & 56.0 & 61.7 & \textbf{24.0} & 26.8 & 31.0 & 33.6 & \textbf{48.0} & \textbf{52.2} & 20.0 & 21.8 & 84.0 & 87.9 & \textbf{78.0} & 83.9 & 64.0 & 39.8 & \textbf{53.6} & 54.0 & 50.3 \\
\addMethod{CoT} & \textbf{56.0} & 66.0 & 45.0 & 0.0 & \textbf{4.7} & 49.7 & 14.5 & 63.4 & \textbf{13.0} & 42.7 & - & \textbf{57.6} & \textbf{64.4} & 23.9 & \textbf{28.2} & \textbf{39.0} & \textbf{41.6} & 46.0 & 50.2 & \textbf{28.0} & \textbf{30.6} & 73.0 & 79.5 & 57.0 & 68.9 & 55.0 & 41.8 & 51.9 & 52.3 & 49.4 \\
\addMethod{FS} & 53.0 & 65.0 & 43.0 & 3.5 & 4.0 & 50.2 & \textbf{15.5} & \textbf{65.8} & 9.0 & 35.0 & \textbf{60.2} & 55.0 & 64.3 & 23.0 & 25.0 & 31.0 & 33.6 & 47.0 & 50.6 & 21.0 & 22.5 & 82.0 & 87.0 & \textbf{78.0} & \textbf{84.5} & 65.0 & 41.1 & 52.8 & \textbf{54.1} & \textbf{50.5} \\
\addMethod{FS CoT} & 54.0 & \textbf{68.0} & \textbf{46.0} & 3.5 & 4.0 & 50.7 & 13.0 & 64.0 & 11.0 & \textbf{43.7} & \textbf{-} & 54.0 & 59.4 & 19.0 & 21.2 & 34.0 & 36.6 & 45.0 & 49.2 & 20.0 & 21.7 & \textbf{89.0} & \textbf{93.8} & 72.0 & 79.7 & \textbf{66.0} & \textbf{42.9} & 52.8 & 53.5 & \textbf{50.5} \\
\midrule

Mistral$_{7b}$ & 47.0 & 50.0 & \textbf{43.0} & \textbf{26.5} & 11.3 & 49.8 & 15.0 & 58.8 & 6.0 & 23.2 & 58.3 & 7.0 & 28.2 & 5.0 & 21.4 & 4.0 & 24.3 & 7.0 & 22.3 & 4.0 & 21.7 & 2.0 & 39.6 & 1.0 & 31.6 & 51.0 & \textbf{41.6} & 47.5 & 30.0 & 37.3 \\
\addMethod{CoT} & 38.0 & 56.0 & 35.0 & 16.5 & \textbf{13.3} & 36.6 & 14.5 & 49.3 & 8.0 & 19.3 & - & 13.0 & 31.3 & 11.0 & 22.4 & 8.0 & 21.1 & \textbf{14.0} & \textbf{24.9} & \textbf{12.0} & \textbf{25.6} & 5.0 & 34.0 & 4.0 & 31.2 & \textbf{61.0} & 36.4 & 35.1 & 31.4 & 33.5 \\
\addMethod{FS} & \textbf{51.0} & 57.0 & 35.0 & 18.0 & 12.0 & \textbf{55.8} & \textbf{25.5} & \textbf{71.0} & \textbf{13.0} & \textbf{52.9} & \textbf{63.0} & \textbf{24.0} & \textbf{43.5} & \textbf{12.0} & 23.9 & 4.0 & 21.3 & 7.0 & 21.2 & 7.0 & 23.0 & 23.0 & 48.9 & 23.0 & 44.9 & 57.0 & 40.3 & \textbf{60.7} & 35.5 & \textbf{43.0} \\
\addMethod{FS CoT} & 29.0 & \textbf{62.0} & 32.0 & 28.5 & 9.3 & 46.6 & 14.5 & 49.2 & 8.0 & 34.4 & - & 13.0 & 33.7 & 11.0 & \textbf{25.3} & \textbf{9.0} & \textbf{26.4} & 9.0 & 24.3 & 9.0 & 20.3 & \textbf{68.0} & \textbf{78.6} & \textbf{42.0} & \textbf{57.4} & 50.0 & 37.9 & 43.4 & \textbf{39.5} & 39.8 \\
\midrule

ChatGLM3$_{6b}$ & \textbf{38.0} & 50.0 & 34.0 & \textbf{2.0} & 3.0 & 34.1 & 7.0 & 43.6 & 14.0 & 56.7 & 38.9 & 20.0 & 41.2 & 14.0 & \textbf{31.7} & 25.0 & 33.8 & 17.0 & 26.0 & 24.0 & 32.2 & 42.0 & 57.0 & 30.0 & 54.0 & 50.0 & \textbf{31.0} & 43.3 & 40.7 & 39.0 \\
\addMethod{CoT} & 27.0 & 49.0 & \textbf{37.0} & 0.0 & 1.0 & 24.8 & 3.0 & 37.1 & 10.0 & 44.8 & - & 24.0 & 41.7 & 10.0 & 25.4 & \textbf{27.0} & 34.6 & 22.0 & 28.1 & 35.0 & 41.2 & 28.0 & 44.5 & 27.0 & 52.0 & 48.0 & 28.3 & 35.6 & 39.4 & 35.7 \\
\addMethod{FS} & 37.0 & 52.0 & 30.0 & \textbf{0.0} & \textbf{10.0} & \textbf{53.0} & \textbf{9.0} & \textbf{52.9} & \textbf{19.0} & \textbf{54.7} & \textbf{50.1} & 11.0 & 19.6 & 13.0 & 17.0 & 25.0 & \textbf{35.0} & \textbf{24.0} & \textbf{30.4} & 20.0 & 25.6 & 27.0 & 46.5 & 33.0 & 53.3 & \textbf{54.0} & 29.8 & \textbf{52.7} & 35.2 & 38.2 \\
\addMethod{FS CoT} & 32.0 & \textbf{66.0} & 31.0 & 0.0 & 4.0 & 34.8 & 8.0 & 43.6 & 11.0 & 44.0 & - & \textbf{25.0} & \textbf{43.3} & \textbf{19.0} & 27.5 & 23.0 & 30.2 & 19.0 & 25.7 & \textbf{36.0} & \textbf{43.0} & \textbf{50.0} & \textbf{62.8} & \textbf{34.0} & \textbf{56.0} & 48.0 & 32.3 & 40.8 & \textbf{42.1} & \textbf{39.2} \\
\midrule

\rowcolor{Gray}
LLaMA2-Base$_{70b}$ & \textbf{55.0} & 61.0 & 37.0 & \textbf{82.0} & \textbf{40.0} & \textbf{67.4} & \textbf{59.0} & \textbf{85.3} & \textbf{63.0} & \textbf{82.7} & \textbf{74.9} & \textbf{57.0} & \textbf{66.7} & \textbf{36.0} & \textbf{48.3} & \textbf{52.0} & \textbf{61.4} & 35.0 & 42.5 & 25.0 & 33.8 & 78.0 & 85.2 & \textbf{80.0} & \textbf{85.4} & 61.0 & 58.8 & \textbf{77.6} & 60.5 & \textbf{64.4} \\
\rowcolor{Gray}
\addMethod{CoT} & 52.0 & \textbf{73.0} & \textbf{39.0} & 79.5 & 32.6 & 62.3 & 47.0 & 79.1 & 25.0 & 61.1 &  & 55.0 & 64.3 & 34.0 & 43.0 & 50.0 & 57.7 & \textbf{37.0} & \textbf{45.2} & \textbf{44.0} & \textbf{53.1} & \textbf{82.0} & \textbf{87.5} & 76.0 & 81.6 & \textbf{67.0} & \textbf{60.9} & 67.5 & \textbf{62.4} & 63.0 \\
\midrule

\rowcolor{Gray}
LLaMA2-Base$_{13b}$ & \textbf{50.0} & 54.0 & 30.0 & 29.5 & 19.3 & 53.3 & \textbf{26.5} & 66.0 & \textbf{20.0} & \textbf{55.6} & \textbf{64.8} & \textbf{48.0} & 59.3 & 34.0 & 48.6 & 41.0 & 49.6 & \textbf{38.0} & 43.4 & 34.0 & 37.5 & 68.0 & \textbf{78.7} & 49.0 & 62.7 & \textbf{58.0} & 40.9 & \textbf{59.9} & 54.7 & 52.6 \\
\rowcolor{Gray}
\addMethod{CoT} & 40.0 & \textbf{61.0} & \textbf{37.0} & 52.0 & \textbf{25.3} & \textbf{59.3} & 26.0 & \textbf{68.8} & 11.0 & 40.8 & - & 46.0 & \textbf{59.4} & \textbf{37.0} & \textbf{49.1} & \textbf{49.0} & \textbf{58.4} & 34.0 & \textbf{43.8} & \textbf{38.0} & \textbf{44.1} & \textbf{70.0} & 78.0 & \textbf{55.0} & \textbf{68.2} & \textbf{58.0} & \textbf{47.5} & 56.3 & \textbf{57.4} & \textbf{54.5} \\
\midrule

\rowcolor{Gray}
LLaMA2-Base$_{7b}$ & 26.0 & 50.0 & 30.0 & 20.0 & 19.3 & 54.5 & 20.0 & 59.6 & 15.0 & \textbf{45.2} & \textbf{62.4} & \textbf{44.0} & \textbf{54.4} & \textbf{30.0} & \textbf{45.3} & \textbf{42.0} & 49.8 & \textbf{34.0} & \textbf{41.9} & 30.0 & 35.8 & \textbf{50.0} & \textbf{64.0} & 36.0 & 53.3 & 49.0 & 31.5 & \textbf{55.4} & 49.2 & 46.3 \\
\rowcolor{Gray}
\addMethod{CoT} & \textbf{37.0} & \textbf{52.0} & \textbf{36.0} & 25.5 & \textbf{21.3} & \textbf{56.9} & \textbf{26.5} & \textbf{67.0} & \textbf{16.0} & 41.9 & - & 32.0 & 45.6 & 27.0 & 36.1 & 41.0 & \textbf{50.9} & 30.0 & 38.0 & \textbf{51.0} & \textbf{57.3} & 45.0 & 59.7 & \textbf{37.0} & \textbf{57.7} & \textbf{50.0} & \textbf{37.6} & 55.3 & \textbf{49.4} & \textbf{47.4} \\
\midrule

\rowcolor{Gray}
Baichuan2-Base$_{13b}$ & 38.0 & 48.0 & 33.0 & 42.5 & 20.7 & 54.8 & \textbf{42.5} & \textbf{73.0} & 11.0 & \textbf{45.7} & \textbf{64.9} & \textbf{50.0} & \textbf{59.4} & \textbf{40.0} & \textbf{54.2} & \textbf{42.0} & \textbf{52.7} & 31.0 & 38.0 & 13.0 & 21.4 & 68.0 & 77.3 & \textbf{50.0} & 63.5 & 54.0 & 40.4 & \textbf{59.6} & 52.6 & 51.3 \\
\rowcolor{Gray}
\addMethod{CoT} & \textbf{50.0} & \textbf{56.0} & \textbf{34.0} & 47.0 & \textbf{29.3} & \textbf{62.0} & 22.5 & 69.3 & \textbf{12.0} & 43.8 & - & 46.0 & 58.2 & 39.0 & 49.6 & 39.0 & 49.8 & \textbf{34.0} & \textbf{40.1} & \textbf{37.0} & \textbf{45.6} & \textbf{73.0} & \textbf{81.3} & 46.0 & \textbf{65.6} & \textbf{60.0} & \textbf{46.8} & 58.4 & \textbf{56.3} & \textbf{54.2} \\
\midrule

\rowcolor{Gray}
Baichuan2-Base$_{7b}$ & 27.0 & \textbf{66.0} & \textbf{41.0} & 32.5 & \textbf{28.0} & \textbf{59.8} & \textbf{34.5} & 69.4 & 5.0 & 34.3 & \textbf{59.8} & 40.0 & \textbf{53.8} & \textbf{35.0} & \textbf{50.2} & \textbf{41.0} & \textbf{49.6} & \textbf{33.0} & \textbf{38.5} & 18.0 & 22.9 & \textbf{49.0} & \textbf{65.9} & 34.0 & 51.0 & \textbf{55.0} & \textbf{41.6} & 55.8 & 48.4 & \textbf{48.5} \\
\rowcolor{Gray}
\addMethod{CoT} & \textbf{30.0} & 56.0 & 34.0 & 34.0 & 23.3 & 57.0 & 33.0 & \textbf{69.5} & \textbf{12.0} & \textbf{44.5} & \textbf{-} & \textbf{41.0} & 51.2 & 31.0 & 40.7 & 38.0 & 46.4 & 26.0 & 32.6 & \textbf{41.7} & \textbf{46.3} & 46.0 & 61.5 & \textbf{43.8} & \textbf{64.1} & 53.0 & 38.5 & \textbf{57.0} & \textbf{49.5} & 48.1 \\
\midrule

\rowcolor{Gray}
Mistral-Base$_{7b}$ & 48.0 & 53.0 & \textbf{38.0} & 41.0 & \textbf{34.0} & \textbf{61.8} & \textbf{42.5} & \textbf{76.2} & \textbf{35.0} & \textbf{61.8} & \textbf{58.3} & 43.0 & 55.9 & \textbf{30.0} & 45.3 & 37.0 & 49.4 & \textbf{38.0} & 47.8 & \textbf{37.0} & \textbf{45.5} & \textbf{68.0} & \textbf{76.7} & \textbf{64.0} & \textbf{74.8} & 53.0 & 45.0 & \textbf{64.5} & \textbf{56.1} & \textbf{55.4} \\
\rowcolor{Gray}
\addMethod{CoT} & \textbf{57.0} & \textbf{63.0} & 35.0 & 54.0 & 30.0 & \textbf{61.8} & 42.0 & 45.7 & 29.0 & 57.3 & - & \textbf{51.0} & \textbf{60.4} & \textbf{30.0} & \textbf{46.2} & \textbf{48.0} & \textbf{57.2} & 37.0 & \textbf{47.9} & 24.0 & 33.2 & 60.0 & 65.9 & 58.0 & 67.9 & \textbf{57.0} & \textbf{52.3} & 54.9 & 54.5 & 54.0 \\
\midrule

\rowcolor{Gray}
ChatGLM3-Base$_{6b}$ & \textbf{48.0} & \textbf{70.0} & \textbf{32.0} & 35.0 & \textbf{3.3} & \textbf{51.8} & \textbf{13.5} & \textbf{62.6} & \textbf{11.0} & 55.0 & \textbf{61.6} & \textbf{50.0} & \textbf{57.2} & \textbf{24.0} & \textbf{26.3} & \textbf{30.0} & \textbf{35.4} & \textbf{38.0} & \textbf{41.5} & \textbf{22.0} & \textbf{22.5} & 67.0 & \textbf{76.4} & 35.0 & \textbf{55.9} & \textbf{58.0} & 46.3 & 57.8 & \textbf{46.7} & \textbf{49.3} \\
\rowcolor{Gray}
\addMethod{CoT} & \textbf{47.0} & \textbf{68.0} & \textbf{32.0} & 46.0 & \textbf{8.7} & \textbf{53.9} & \textbf{15.5} & \textbf{64.3} & \textbf{13.0} & 56.5 &  & 45.0 & 52.5 & 23.0 & 24.5 & 30.0 & \textbf{35.0} & 37.0 & 40.2 & \textbf{22.0} & \textbf{22.5} & \textbf{72.0} & \textbf{79.4} & \textbf{42.0} & \textbf{60.3} & 54.0 & \textbf{48.3} & \textbf{58.2} & 46.1 & 49.1\\
\midrule

\bottomrule

\caption{Full results of TimeBench. Aligned models are under zero-shot setting by default. Methods with \dag ~are base models without alignment, under few-shot setting, thus incomparable with other methods. We consider human performance as 100 points and normalize models' results accordingly.}
\label{tab:full_results}
\end{longtable}
\end{landscape}
\clearpage
\twocolumn

\begin{figure*}[h]
    \centering
    \footnotesize
    \demonstrationfigure{\textsc{DurationQA, MCTACO}}{
        Answer the following question, select all the possible correct options, and each question has at least one correct option. \\
         Context: \{\} \\
        Question: \{\} \\
        Options: \{\} \\
        Answer:\\
    }
    \demonstrationfigure{\textsc{TimeDial}}{
        There is a two-person dialogue with several options. \\
        Choose all appropriate options to substitute the <mask> in the dialogue, and each question has at least one correct option. \\
        Dialogue: \{\} \\
        Options: \{\} \\
        Answer:
    }
    \demonstrationfigure{\textsc{TRACIE}}{
        Read the following story and hypothesis, determine whether the hypothesis can be inferred from the story. \\
        You need to understand the implicit temporal relationships between events to make judgments.\\
        Story: \{\} \\
        Hypothesis: \{\} \\
        Options: A. Entailment  B. Contradiction \\
        Answer: \\
    }
    \demonstrationfigure{\textsc{SituatedGen}}{
        Generate a pair of contrastive sentences with the given set of keywords. \\
        Keywords: \{\}
    }
    \demonstrationfigure{\textsc{Date Arithmetic}}{
        Question: \{\}? Answer:
    }
    \demonstrationfigure{\textsc{TimeQA}}{
        I will give you a question with context. \\
        You need to answer my question based on the context.\\
        If you can infer the answer from the context, then output your answer. Otherwise, if there is no answer, output [unanswerable].\\
        Context: \{\} \\
        Question: \{\} \\
        Answer:
    }
    \demonstrationfigure{\textsc{TempReason}}{
        I will give you a question with context. \\
        You need to answer my question based on the context.\\
        Context: \{\} \\
        Question: \{\} \\
        Answer:
    }
    \demonstrationfigure{\textsc{MenatQA}}{
        Get answers for the question based on the contxt, where answers derived from substrings in the context or categorized as [unanswerable].\\
        Context: \{\} \\
        Question: \{\} \\
        Answer: \\
    }
    \demonstrationfigure{\textsc{TimeX-NLI}}{
        Read the following statements about time and determine if the hypothesis can be inferred from the premise. \\
        Premise: \{\} \\
        Hypothesis: \{\} \\
        Options: A. Entailment  B. Contradiction  C. Neutral \\
        Answer:
    }

    \caption{Zeroshot instructions and input formats.}
    \label{fig:prompt-instruction}
\end{figure*}
\begin{figure*}[h]
    \centering
    \footnotesize
    \demonstrationfigure{CoT Demonstration of \textsc{TimeX-NLI} (3-shot, order)}{
        Answer the following question, select all the possible correct options, and each question has at least one correct option. \newline
        
        \textbf{Premise}: On Wednesday, they got married.\\
        \textbf{Hypothesis}: Before Friday, they got married.\\
        \textbf{Options}: A. Entailment  B. Contradiction  C. Neutral\\
        \textbf{Answer}: Wednesday is before Friday. As a result, we can infer that if something happens on Wednesday, it definitely happens before Friday. Therefore, the answer is A. Entailment.\newline
        
        \textbf{Premise}: We went to Disneyland on Monday.\\
        \textbf{Hypothesis}: We went to Disneyland after Wednesday.\\
        \textbf{Options}: A. Entailment  B. Contradiction  C. Neutral\\
        \textbf{Answer}: Monday is before Wednesday. As a result, We can infer that if something happens on Monday, it definitely can not happen after Wednesday. Therefore, the answer is B. Contradiction.\newline
        
        \textbf{Premise}: The failing company issued major layoffs after Tuesday.\\
        \textbf{Hypothesis}: The failing company issued major layoffs after Thursday.\\
        \textbf{Options}: A. Entailment  B. Contradiction  C. Neutral\\
        \textbf{Answer}: Tuesday is before Thursday. If something happened after Tuesday, we cannot be certain whether it occurred after Thursday. Therefore, the answer is C. Neutral.     
    }
    
    \caption{Chain-of-Thought demonstrations of TimeX-NLI (s1-order).}
    \label{fig:prompt-demos-timexnli}
\end{figure*}
\begin{figure*}[h]
    \centering
    \footnotesize
    \demonstrationfigure{CoT Demonstration of \textsc{Date Arithmetic} (4-shot)}{
        \textbf{Question}: What is the time 4 year and 1 month after Apr, 2000?\\
        \textbf{Answer}: First, 4 years after 2000 is 2004. Next, 1 month after April is May. Therefore, 4 year and 1 month after Apr, 2000 is May, 2004.\newline
        
        \textbf{Question}: What is the time 3 year and 4 month before Jun, 1840?\\
        \textbf{Answer}: First, subtracting 3 years from 1840 gives 1837. Next, subtracting 4 months from June gives February. Therefore, 3 year and 4 month before Jun, 1840 is Feb, 1837.\newline
        
        \textbf{Question}: What is the time 7 year and 11 month after Feb, 1819?\\
        \textbf{Answer}: First, 7 years after 1819 is 1826. Next, 11 months after February is January of the next year. Therefore, 7 years and 11 months after Feb, 1819 is Jan, 1827.\newline
        
        \textbf{Question}: What is the time 6 year and 9 month before Jan, 1234?\\
        \textbf{Answer}: First, subtracting 6 years from 1234 gives 1228. Next, subtracting 9 months from January gives April of the previous year. Therefore, 6 year and 9 month before Jan, 1234 is Apr, 1227.
    }

    \caption{Chain-of-Thought demonstrations of Date Arithmetic.}
    \label{fig:prompt-demos-arith}
\end{figure*}
\begin{figure*}[h]
    \centering
    \footnotesize
    \demonstrationfigure{CoT Demonstration of \textsc{TRACIE} (4-shot)}{
        Read the following story and hypothesis, determine whether the hypothesis can be inferred from the story.\\
        You need to understand the implicit temporal relationships between events to make judgments \newline
        
        ......\newline
        
        \textbf{Story}: Joe was a police officer. Joe was patrolling the streets of the city in his cruiser. \"Suddenly, Joe was alerted to a crime happening near him by dispatch.\" Joe responded to the scene and found a bank robber fleeing on foot. Joe arrested the criminal and was promoted.\\
        Hypothesis: Joe put on his police uniform. starts after Joe arrest the criminal\\
        \textbf{Options}: A. Entailment  B. Contradiction\\
        \textbf{Answer}: From the story we know Joe was patrolling. In the work state, Joe has already put on the police uniform. So we can infer that Joe put on his police uniform before arresting the criminal. This conflicts with hypothesis. Therefore, the answer is B. Contradiction.
    }
    
    \caption{Chain-of-Thought demonstrations of TRACIE.}
    \label{fig:prompt-demos-tracie}
\end{figure*}
\begin{figure*}[h]
    \centering
    \footnotesize
    \demonstrationfigure{CoT Demonstration of \textsc{DurationQA} (4-shot)}{
        Answer the following question, select all the possible correct options, and each question has at least one correct option.\newline
        
        ......\newline
        
        \textbf{Context}: actually i have an project on it so please give me as much as you have information about migratory birds in punjab\\
        \textbf{Question}: How long did it take for them to have information about migratory birds in punjab?\\
        \textbf{Options}: A. several months  B. 12 weeks  C. a few minutes  D. almost instantly\\
        \textbf{Answer}: This is a conversation scenario. In the conversation, providing relevant information about migratory birds in punjab to him is in real-time and takes very little time. Therefore, the answer is C. a few minutes, D. almost instantly.\newline
        
        \textbf{Context}: Hope she stops laying eggs because she will get really skinny !\\
        \textbf{Question}: How long did it take for her to lay eggs?\\
        \textbf{Options}: A. 1 week  B. 22 hours  C. 2 years  D. 4 years\\
        \textbf{Answer}: According to commonsense knowledge, the time it takes for birds to lay eggs typically varies from one day to several days. Therefore, the answer is A. 1 week, B. 22 hours.
    }

    \caption{Chain-of-Thought demonstrations of DurationQA.}
    \label{fig:prompt-demos-durationqa}
\end{figure*}

\begin{figure*}[t]
    \centering
    \footnotesize
    \demonstrationfigure{CoT Demonstration of \textsc{MCTACO} (4-shot)}{
        Answer the following question, select all the possible correct options, and each question has at least one correct option.\newline
        
        ......\newline
        
        \textbf{Context}: She ordered the tastiest kind of each vegetable and the prettiest kind of each flower.\\
        \textbf{Question}: How often does she order vegetables and flowers?\\
        \textbf{Options}: A. once a second  B. three days a week  C. every 10 centuries  D. once a week\\
        \textbf{Answer}: According to commonsense knowledge, ordering vegetables and flowers typically happens on a regular basis, usually every few days. Therefore, the answer is B. three days a week, D. once a week.\newline
        
        \textbf{Context}: Wallace, 38, called Gastonia home from the age of 8 until she graduated from Hunter Huss High School in 1983.\\
        \textbf{Question}: When did Wallace wake up for high school?\\
        \textbf{Options}: A. at 6 am  B. at 1 am  C. 7:00 AM  D. at 6 pm\\
        \textbf{Answer}: According to commonsense knowledge, waking up for high school typically happens in the morning, usually between 6 AM and 8 AM. Therefore, the answer is A. at 6 am, C. 7:00 AM.
    }
    
    \caption{Chain-of-Thought demonstrations of MCTACO.}
    \label{fig:prompt-demos-mctaco}
\end{figure*}

\begin{figure*}[t]
    \centering
    \footnotesize
    \demonstrationfigure{CoT Demonstration of \textsc{TimeDial} (4-shot)}{
        There is a two-person dialogue with several options.\\
        Choose all appropriate options to substitute the <mask> in the dialogue, and each question has at least one correct option.\newline
        
        ......\newline
        
        \textbf{Dialogue}: \\
        A:What schools have you attended ?\\
        B: I finished Young Primary School in 1998 , and entered Xi ' an Middle School that same September . I graduated from there in <MASK> , and that September I entered Wuhan University , where I'm studying now .\\
        A: How do you think the education you have received will contribute to your work in this company ?\\
        B: I think I have a good understanding of fundamentals in the areas your company deals with , and I can go on from here to build up the specific skills and knowledge I need to do my job well .\\
        A: Your graduation thesis was on Medical Application of Laser , right ? What were your conclusions ?\\
        B: Yes . I did some work on that , and I found out some really interesting things about the conductivity of liquid helium . I was sure I had a great discovery until my teacher told me the same discovery already made twenty years ago . I think the most important thing , I learnt though , was the importance of keeping good records .\\
        \textbf{Options}: A. 1998  B. July of 2004  C. March of 2003  D. twenty years ago\\
        \textbf{Answer}: Based on the dialogue, B entered middle school in Sep 1998. According to commonsense knowledge, it usually takes around 6 years from entering middle school to graduating from high school (and entering university). Adding 6 years to 1998 would be 2004, so the answer should be around the year 2004. Therefore, the answer is B. July of 2004, C. March of 2003.
    }
   
    \caption{Chain-of-Thought demonstrations of TimeDial.}
    \label{fig:prompt-demos-timedial}
\end{figure*}
\begin{figure*}[ht]
    \centering
    \footnotesize
    \demonstrationfigure{CoT Demonstration of \textsc{TimeQA, MenatQA} (2-shot, implicit)}{
        I will give you a question with context.\\
        You need to answer my question based on the context.\\
        If you can infer the answer from the context, then output your answer. Otherwise, if there is no answer, output [unanswerable]\newline
        
        ......\newline
        
        \textbf{Context}: Theo-Ben Gurirab Theo-Ben Gurirab ( 23 January 1938 \u2013 14 July 2018 ) was a Namibian politician who served in various senior government positions . He served as the second Prime Minister of Namibia from 28 August 2002 to 20 March 2005 , following the demotion and subsequent resignation of Hage Geingob . Previously he was the countrys first Minister of Foreign Affairs from 1990 to 2002 , and was President of the United Nations General Assembly from 1999 to 2000 . He was Speaker of the National Assembly of Namibia from 2005 to 2015 , when he was replaced by Peter Katjavivi . Gurirab ultimately resigned from politics in 2015 . Death . Gurirab died at a Windhoek hospital on 14 July 2018 of natural causes . He is buried at Heroes Acre .\\        
        \textbf{Question}: Theo-Ben Gurirab took which position after Jan 2007?\\
        \textbf{Answer}: Based on the context, we can summarize the following facts: Theo-Ben Gurirab served as second Prime Minister of Namibia from August 2002 to March 2005. Prior to that, he was the countrys first Minister of Foreign Affairs from 1990 to 2002 and and was President of the United Nations General Assembly from 1999 to 2000. From 2005 to 2015, he held the position of Speaker of the National Assembly of Namibia. He resigned from politics in 2015 and passed away in July 2018. According to the aforementioned facts, he took the position of Speaker of the National Assembly of Namibia in January 2007. Therefore, the answer is Speaker of the National Assembly of Namibia.
    }
   
    \caption{Chain-of-Thought demonstrations of TimeQA, MenatQA, implicit reasoning.}
    \label{fig:prompt-demos-timeqa}
\end{figure*}

\begin{figure*}[t]
    \centering
    \footnotesize
    \demonstrationfigure{CoT Demonstration of \textsc{TempReason} (4-shot, event-time)}{
        I will give you a question with context.\\
        You need to answer my question based on the context.\newline
        
        ......\newline
        
        \textbf{Context (facts)}: Gian Piero Gasperini is the head coach of Atalanta B.C. from Jun, 2016 to Dec, 2022. \\ Edoardo Reja is the head coach of Atalanta B.C. from Mar, 2015 to Jun, 2016. \\ Stefano Colantuono is the head coach of Atalanta B.C. from Jun, 2010 to Mar, 2015. \\ Bortolo Mutti is the head coach of Atalanta B.C. from Jan, 2010 to Jun, 2010. \\ Emiliano Mondonico is the head coach of Atalanta B.C. from Jul, 1987 to Jun, 1990. \\ Marcello Lippi is the head coach of Atalanta B.C. from Jul, 1992 to Jun, 1993. \\ Angelo Gregucci is the head coach of Atalanta B.C. from Jul, 2009 to Sep, 2009. \\ Luigi Delneri is the head coach of Atalanta B.C. from Jul, 2007 to Jun, 2009. \\ Ottavio Bianchi is the head coach of Atalanta B.C. from Jul, 1981 to Jun, 1983. \\ Antonio Conte is the head coach of Atalanta B.C. from Sep, 2009 to Jan, 2010. \\ Nedo Sonetti is the head coach of Atalanta B.C. from Jul, 1983 to Jun, 1987. \\ Valter Bonacina is the head coach of Atalanta B.C. from Jan, 2010 to Jan, 2010.\\
        \textbf{Question}: Who was the head coach of the team Atalanta B.C. in Feb, 2016?\\
        \textbf{Answer}: According to the context,  Edoardo Reja was the head coach of Atalanta B.C. from Mar, 2015 to Jun, 2016. In Feb 2016, the head coach of the team Atalanta B.C. is Edoardo Reja. Therefore, the answer is Edoardo Reja.
    }
    
    \caption{Chain-of-Thought demonstrations of TempReason, event-time reasoning.}
    \label{fig:prompt-demos-tempreason-et}
\end{figure*}

\begin{figure*}[t]
    \centering
    \footnotesize
    \demonstrationfigure{CoT Demonstration of \textsc{TempReason} (4-shot, event-event)}{
        I will give you a question with context.\\
        You need to answer my question based on the context.\newline
        
        ......\newline
        
        \textbf{Context (facts)}: Nicholas Macpherson holds the position of Member of the House of Lords from Oct, 2016 to Dec, 2022. \\Nicholas Macpherson holds the position of Principal Private Secretary to the Chancellor of the Exchequer from Jan, 1993 to Jan, 1997. \\Nicholas Macpherson holds the position of Permanent Secretary to the Treasury from Aug, 2005 to Jan, 2016.\\
        \textbf{Question}: Which position did Nicholas Macpherson hold before Member of the House of Lords?\\
        \textbf{Answer}: According to the context, Nicholas Macpherson holds the position of Permanent Secretary to the Treasury from Aug, 2005 to Jan, 2016. Afterthat, Nicholas Macpherson holds the position of Member of the House of Lords from Oct, 2016 to Dec, 2022. Nicholas Macpherson hold the position of Permanent Secretary to the Treasury before Member of the House of Lords. Therefore, the answer is Permanent Secretary to the Treasury."
    }
   
    \caption{Chain-of-Thought demonstrations of TempReason, event-event reasoning.}
    \label{fig:prompt-demos-tempreason-ee}
\end{figure*}

\end{document}